\documentclass{article} 
\usepackage{iclr2027_conference,times}

\usepackage{amsmath,amsfonts,bm}

\def\eqref#1{equation~\ref{#1}}

\def\1{\bm{1}}

\def\eps{{\epsilon}}

\def\va{{\bm{a}}}
\def\vb{{\bm{b}}}

\def\ve{{\bm{e}}}

\def\vk{{\bm{k}}}
\def\vl{{\bm{l}}}

\def\vn{{\bm{n}}}
\def\vo{{\bm{o}}}
\def\vp{{\bm{p}}}
\def\vq{{\bm{q}}}
\def\vr{{\bm{r}}}
\def\vs{{\bm{s}}}

\def\vu{{\bm{u}}}
\def\vv{{\bm{v}}}

\def\vx{{\bm{x}}}

\def\vz{{\bm{z}}}

\def\mM{{\bm{M}}}

\def\mR{{\bm{R}}}

\DeclareMathAlphabet{\mathsfit}{\encodingdefault}{\sfdefault}{m}{sl}
\SetMathAlphabet{\mathsfit}{bold}{\encodingdefault}{\sfdefault}{bx}{n}

\def\gL{{\mathcal{L}}}

\def\gN{{\mathcal{N}}}
\def\gO{{\mathcal{O}}}

\def\gU{{\mathcal{U}}}

\usepackage{amsmath}
\usepackage{amssymb}
\usepackage{graphicx}
\usepackage{booktabs}
\usepackage{multirow}
\usepackage{xcolor}
\usepackage{hyperref}
\usepackage{url}
\usepackage[letterspace=80]{microtype}

\usepackage{tikz}
\usetikzlibrary{shapes.geometric, arrows.meta, positioning, fit, calc, backgrounds}
\usepackage{bm}

\definecolor{panelborder}{HTML}{CCCCCC}
\definecolor{panelbg}{HTML}{F9F9F9}
\definecolor{navy}{HTML}{1E3A8A}
\definecolor{txtgray}{HTML}{666666}
\definecolor{tokblue}{HTML}{2563EB}
\definecolor{sampred}{HTML}{DC2626}
\definecolor{temporal}{HTML}{D97706}
\definecolor{teal}{HTML}{0D9488}
\definecolor{purple}{HTML}{9333EA}
\definecolor{orange}{HTML}{EA580C}
\definecolor{green}{HTML}{16A34A}

\title{Learning Incompressible 3D Flows from Sparse Samples: A Set Transformer Approach to Divergence-Free Kernel Fitting}

\author{
Ramzi Idoughi\textsuperscript{1}\thanks{Corresponding author: \texttt{ramzi.idoughi@kaust.edu.sa}} ,
Radouan Boukharfane\textsuperscript{2},
Peter Wonka\textsuperscript{1} \vspace{0.4cm} \\
\textsuperscript{1}King Abdullah University of Science and Technology (KAUST), Thuwal, Saudi Arabia\\
\textsuperscript{2}Mohammed VI Polytechnic University (UM6P), Benguerir, Morocco \\
}

\DeclareMathOperator{\EPE}{EPE}

\iclrfinalcopy

\begin{document}
\maketitle
\providecommand{\eps}{\varepsilon}
\providecommand{\dfk}{\textsc{dfk}}
\providecommand{\Rbb}{\mathbb{R}}

\definecolor{dfkpos}{HTML}{1F3F6E}
\definecolor{dfkvel}{HTML}{B85C3E}
\definecolor{dfkconf}{HTML}{6E6E6E}
\definecolor{dfktime}{HTML}{D95F2B}
\definecolor{dfkcenter}{HTML}{2E6DA4}
\definecolor{dfkradius}{HTML}{1A8A8A}
\definecolor{dfkaspect}{HTML}{B37D00}
\definecolor{dfkorient}{HTML}{7A4A20}
\definecolor{dfkweight}{HTML}{C41E7A}
\definecolor{dfkmean}{HTML}{7B3FBF}
\definecolor{dfkfield}{HTML}{1F9A5A}
\definecolor{dfkBlack}{HTML}{000000}

\newif\ifdfkcolors \dfkcolorstrue
\newcommand{\dfkcolorsoff}{\dfkcolorsfalse}
\newcommand{\dfkcolorson}{\dfkcolorstrue}

\newcommand{\dfkc}[2]{{\ifdfkcolors\color{#1}\fi#2}}

\providecommand{\vomega}{{\bm{\omega}}}

\newcommand{\obsp}{\dfkc{dfkpos}{\vp}}
\newcommand{\obspm}{\dfkc{dfkpos}{\vp_m}}
\newcommand{\obsu}{\dfkc{dfkvel}{\vu}}
\newcommand{\obsum}{\dfkc{dfkvel}{\vu_m}}
\newcommand{\obsc}{\dfkc{dfkconf}{c}}
\newcommand{\obscm}{\dfkc{dfkconf}{c_m}}
\newcommand{\tvar}{\dfkc{dfkBlack}{t}}

\newcommand{\kcen}{\dfkc{dfkcenter}{\vx}}
\newcommand{\kcenk}{\dfkc{dfkcenter}{\vx_k}}
\newcommand{\kceninit}{\dfkc{dfkcenter}{\vx_k^{0}}}
\newcommand{\kcendel}{\dfkc{dfkcenter}{\Delta\vx_k}}
\newcommand{\kcent}{\dfkc{dfkcenter}{\vx_k(\tvar)}}
\newcommand{\kadv}{\dfkc{dfkcenter}{\vv^{\mathrm{adv}}_k}}
\newcommand{\hcen}{\dfkc{dfkcenter}{h^{c}_k}}

\newcommand{\krad}{\dfkc{dfkradius}{\sigma}}
\newcommand{\kradk}{\dfkc{dfkradius}{\sigma_k}}
\newcommand{\kradfix}{\dfkc{dfkradius}{\sigma_{\mathrm{fixed}}}}
\newcommand{\hrad}{\dfkc{dfkradius}{h^{r}_k}}

\newcommand{\krot}{\dfkc{dfkorient}{\mR}}
\newcommand{\krotk}{\dfkc{dfkorient}{\mR_k}}
\newcommand{\kmap}{\dfkc{dfkorient}{\mM_k}}

\newcommand{\kasp}{\dfkc{dfkaspect}{\va}}
\newcommand{\kaspk}{\dfkc{dfkaspect}{\va_k}}
\newcommand{\kaspperp}{\dfkc{dfkaspect}{a_{\perp}}}
\newcommand{\kasplong}{\dfkc{dfkaspect}{a_{\mathrm{long}}}}

\newcommand{\kwgt}{\dfkc{dfkweight}{\vomega}}
\newcommand{\kwgtk}{\dfkc{dfkweight}{\vomega_k}}
\newcommand{\kwgtkt}{\dfkc{dfkweight}{\vomega_k}(\tvar)}
\newcommand{\kfdc}{\dfkc{dfkweight}{\va_k^{0}}}
\newcommand{\kfcos}{\dfkc{dfkweight}{\va_k^{f}}}
\newcommand{\kfsin}{\dfkc{dfkweight}{\vb_k^{f}}}

\newcommand{\vmean}{\dfkc{dfkmean}{\bar{\vv}}}
\newcommand{\vmeant}{\dfkc{dfkmean}{\bar{\vv}}(\tvar)}

\newcommand{\vfld}{\dfkc{dfkfield}{\vv_{\Theta}}}

\newcommand{\vstar}{{\vv^{\ast}}}
\begin{abstract}
Recovering divergence-free velocity fields from sparse and corrupted observations requires reconciling physical consistency, expressive representations, and robustness to missing data. We introduce $\textbf{DFK-Former}$, a set-to-field transformer that reconstructs divergence-free velocity fields from sparse, noisy, and irregularly sampled 3D observations. The model is inherently invariant to the number and ordering of input samples and predicts the parameters of a divergence-free kernel representation, producing continuous, mesh-free velocity fields that satisfy incompressibility exactly by construction rather than approximately through soft constraints or projection operators. Unlike existing divergence-free kernel approaches, which estimate kernel coefficients via a separate optimization for each field, DFK-Former predicts these coefficients directly in a single forward pass. This eliminates costly iterative fitting while providing a learned prior that enables accurate reconstruction even when observations are insufficient to uniquely determine the field. To further enhance representational power, we introduce adaptive and anisotropic divergence-free kernels that preserve exact incompressibility guarantees while capturing more complex flow structures. Extensive experiments across varying levels of sparsity, noise, and occlusion, including transfer to real Direct Numerical Simulation (DNS) turbulence data, demonstrate substantial gains in both accuracy and efficiency. DFK-Former reduces endpoint error by up to $4.6~\times$ compared to per-field divergence-free-kernel optimization, while reconstructing each field in $7.5~ms$ instead of $9.8~s$, yielding a speedup of more than three orders of magnitude. These results establish direct, feed-forward prediction of divergence-free kernel parameters as a practical and effective paradigm for physically consistent velocity-field reconstruction from sparse observations.
\end{abstract}
\section{Introduction}
\label{sec:intro}

Reconstructing dense, physically plausible velocity fields is a fundamental challenge in experimental fluid dynamics and simulation-driven graphics, with applications ranging from flow
characterization and numerical-model validation to data-driven simulation. Particle Imaging Velocimetry (PIV) and its variants measure flow by tracking passive
tracers~\citep{adrian2011particle, scarano2013tomographic, xiong2021rainbowpiv, wang2020stereo}. Alternative approaches infer velocity from light-transport
measurements~\citep{atcheson2008time, ji2013reconstructing} or directly from RGB video using physical constraints~\citep{eckert2019scalarflow, zang2020tomofluid}. Sensor-network approaches,
in turn, infer flow fields from sparse spatial observations~\citep{erichson2020shallow, nguyen2026deep}. Regardless of the modality, the measurements are sparse, incomplete, and contaminated by
noise, making recovery of the underlying velocity field a challenging task but ideally suited for machine learning.

In many fluid-flow settings, the underlying velocity field is approximately incompressible and should therefore have zero divergence (i.e., divergence-free). However, in practice, measured observations $\obsum$ deviate from this physical constraint due to noise, occlusions, limited spatial resolution, and interpolation artifacts. As a result, reconstructing the physically admissible flow involves estimating the divergence-free component of the observations, typically through the Leray (Helmholtz--Hodge) projection: $\vstar = P_{\mathrm{\textit{div-free}}}(\obsu)$.
Although straightforward in a densely sampled setting, the problem becomes more challenging when observations are sparse and corrupted.

Recent approaches to divergence-free reconstruction address complementary aspects of this problem. Divergence-Free Kernels (DFKs)~\citep{ni2025representing} represent velocity fields as linear combinations of compactly-supported matrix-valued basis functions obtained by applying differential operators to Wendland polynomial $C^{4}$~\citep{ wendland1995piecewise}. These kernels guaranty exact incompressibility by construction, while yielding mesh-free and continuously queryable velocity representations. However, DFKs are fitted on a per-instance basis. Consequently, every reconstruction requires the solution of a new optimization problem from scratch, incurring a high computational cost without learning prior knowledge about the distribution of physically plausible flows. In contrast, HDNet~\citep{qi2024hdnet} learns a Helmholtz decomposition from data using a convolutional architecture, effectively capturing statistical regularities of flow fields through a learned prior. Yet its divergence-free output is only approximate, relying on discrete grid-based operations without exact guarantees, and the method assumes dense observations on a structured 2D grid. To the best of our knowledge, no existing method unifies learned flow priors with exact divergence-free guarantees for sparse, unstructured 3D observations.

Recent advances in permutation-equivariant architectures, including Set Transformer~\citep{lee2019set}, Perceiver~IO~\citep{jaegle2022perceiver}, and VGGT~\citep{wang2025vggt}, have demonstrated the ability to infer structured geometric representations directly from unordered sets of observations in a single forward pass. Building on this idea, we propose \textbf{DFK-Former}, which couples a transformer-based set encoder with a divergence-free kernel representation. Given an unordered set of observations $\gO = \{(\obspm, \obsum, \obscm)\}_{m=1}^{M}$ consisting of 3D positions, noisy velocity measurements, and confidence scores, \textbf{DFK-Former} predicts the parameters ($\Theta$) of a DFK representation $\vfld \approx \vstar$. The resulting velocity field is exactly divergence-free by construction, remains mesh-free and continuously queryable, and can be reconstructed in a single forward pass without per-instance optimization at inference time.\\
The main contributions of this work are as follows: 
\begin{itemize}
\item We introduce \textbf{DFK-Former}, which combines learned flow priors with exact divergence-free guarantees for sparse, unstructured 3D observations by predicting kernel weights,
locations, and support radii directly, without per-instance inference optimization.
\item We develop anisotropic divergence-free kernels that capture coherent flow structures and improve reconstruction at matched model capacity.
\item We extend the model to unsteady flows using alternating spatial and temporal attention with Fourier-in-time kernels, preserving exact incompressibility at every instant.
\item We establish a comprehensive evaluation framework and demonstrate gains across sparsity, noise, and occlusion, including on real DNS data.
\end{itemize}

\section{Related Work}
\label{sec:related}

\textbf{Flow Reconstruction from Sparse Observations.} 
Recovering dense velocity fields from incomplete and noisy measurements is a central problem in fluid dynamics and scientific machine learning.
In experimental fluid mechanics, PIV and its extensions estimate flow fields by tracking tracer particles across image sequences~\citep{adrian2011particle, scarano2013tomographic, wang2020stereo, xiong2021rainbowpiv}.
Related approaches reconstruct fluid motion from light transport measurements, scalar transport observations, or monocular RGB videos~\citep{atcheson2008time, ji2013reconstructing, eckert2019scalarflow, zang2020tomofluid}. Across modalities, measurements suffer from limited coverage, occlusions, artifacts, and noise.
Classical remedies use interpolation, variational optimization, or data assimilation; recent learning-based methods reconstruct fields from sparse sensors, images, or scattered samples~\citep{erichson2020shallow, nguyen2026deep}. A closely related direction represents physical quantities using continuous coordinate-based neural fields, enabling mesh-free reconstruction from sparse observations~\citep{sitzmann2020implicit, tancik2020fourier}. Applications include Neural Flow Maps~\citep{sahoo2022neural} and Physics-Enhanced Implicit Neural Representations~\citep{shen2025peinr}. However, these methods solve a separate optimization problem for each scene and enforce physical constraints only approximately. In contrast, our approach learns a direct mapping from sparse observations to velocity fields while targeting exact incompressibility.

\textbf{Divergence-Free Representations and Physical Constraints.}
Many learning-based approaches incorporate physical consistency through regularization terms, projection operators, or physics-informed objectives~\citep{raissi2019physics, li2021fourier}. For incompressible flows, several methods encourage divergence-free solutions via soft penalties, projection operators, or vector-potential parameterizations~\citep{raissi2020hidden, cai2021physics, tompson2017accelerating, qi2024hdnet, li2026project, chang2021curl, lyu2024wavelet}. A complementary line of work encodes incompressibility directly in the representation itself, either through divergence-free neural architectures based on differential-form parameterizations~\citep{richterpowell2022neural} or through divergence-free radial basis functions and matrix-valued kernels that guarantee analytically divergence-free vector fields over the continuous domain~\citep{narcowich1994generalized, wendland1995piecewise}. Building on this foundation, Divergence-Free Kernels (DFKs)~\citep{ni2025representing} introduced a compactly supported, continuously queryable representation for incompressible velocity fields with exact physical guarantees. Subsequent studies have explored integrating DFKs into broader fluid reconstruction~\citep{tao2026lagrangiansplats} and simulation~\citep{xing2026ddfks} pipelines. Unlike prior DFK-based approaches, which solve an optimization problem independently for each scene, our method learns to predict kernel parameters directly from observations, combining exact incompressibility with feed-forward reconstruction. 

  \paragraph{Set-Based Geometric Inference and Operator Learning.}
  Learning-based reconstruction replaces expensive per-instance optimization with a predictor trained across instances~\citep{garnelo2018conditional, gordon2018meta, cranmer2020frontier,
  marino2018iterative}, and has proven effective for inverse problems and implicit representations~\citep{siahkoohi2021preconditioned, tancik2021learned}. Since sparse flow measurements form
  unordered sets rather than structured grids, architectures that process them should be permutation-invariant or equivariant by construction. Deep Sets~\citep{zaheer2017deep} established
  this paradigm, and attention-based architectures such as Set Transformer~\citep{lee2019set} and Perceiver IO~\citep{jaegle2022perceiver} extended it to model interactions among set
  elements. Building on these ideas, recent transformers infer continuous fields or 3D structure from irregular observations, spanning scattered-data interpolation and operator
  learning~\citep{ding2024niert, hao2023gnot, wu2024transolver}, as well as feed-forward 3D reconstruction from sparse views~\citep{wang2025vggt}.
  Similarly, operator-learning methods such as DeepONet~\citep{lu2021learning} and Fourier Neural Operators~\citep{li2021fourier} learn mappings between function spaces and have achieved
  strong results in fluid modeling. Several approaches leverage these ideas for sparse-sensor flow reconstruction~\citep{zhao2024recfno, zhang2025operator, dang2026deep}. These methods
  regress field values without building physical constraints into the representation, so incompressibility is satisfied only approximately. In contrast, \textbf{DFK-Former} infers, in one
  forward pass, the parameters of a continuous divergence-free kernel representation with a permutation-equivariant transformer.

\section{Background: Divergence-Free Kernel Fields} 
\label{sec:background} 

\begin{figure}[t]
\centering
\includegraphics[width=0.87\linewidth]{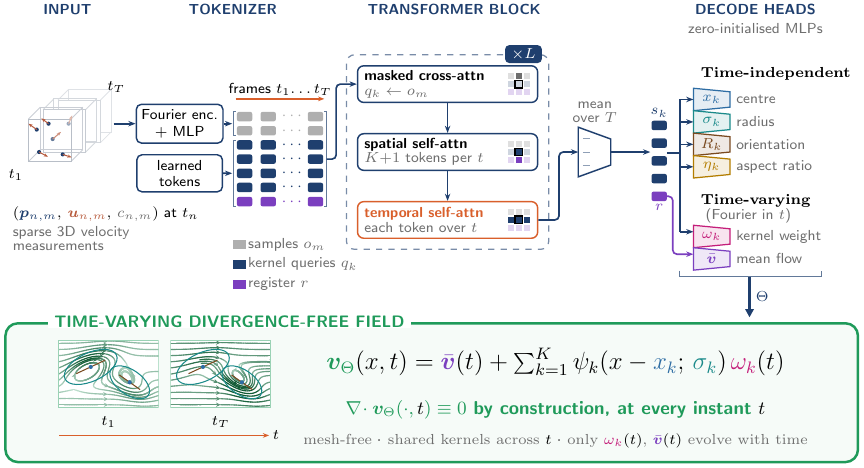}
\caption{\textbf{DFK-Former architecture}. Sparse spatiotemporal velocity observations are encoded into tokens and jointly processed with learned kernel-query and register tokens using a Perceiver-style spatiotemporal transformer. Decoding heads predict the parameters of a divergence-free kernel representation, yielding a continuous time-varying incompressible velocity field by construction.}
\label{fig:pipeline}
\end{figure}

Modeling incompressible flows requires velocity fields ($\vv:\mathbb{R}^3\rightarrow\mathbb{R}^3$) satisfying $\nabla \cdot \vv (\vx) = 0$.
Rather than enforcing incompressibility through divergence penalties or projection, Divergence-Free Kernels (DFKs)~\citep{ni2025representing} guarantee this constraint analytically by construction. The velocity field is expressed as a superposition of matrix-valued divergence-free kernels as follows:
\begin{equation}
    \vv(\vx) = \sum_{k=1}^{K} \psi_{\mathrm{iso}}(\vx-\kcenk;\kradk)\,\kwgtk, 
    \label{eq:dfk-no-mean} 
\end{equation} 
where $\kcenk\in\mathbb{R}^3$ is the kernel center, $\kradk>0$ its support radius, and $\kwgtk\in\mathbb{R}^3$ its vector coefficient, which controls its contribution to the flow. Using the divergence-free operator $\mathcal{D}=-\mathbf{I}\Delta+\nabla\nabla^\top$ \footnote{The derivatives in $\mathcal{D}$ are taken with
  respect to $\vx$.}, we define the isotropic kernel using the normalized Wendland-$\mathcal{C}^4$ potential as for~\citep{ni2025representing}:
  \begin{equation}
  \psi_{\mathrm{iso}}(\vx;\krad)=\mathcal{D}\,\phi\!\left(\frac{\|\vx\|}{\krad}\right),
  \qquad
  \phi(r)=\frac{(\max(1-r,0))^6(35r^2+18r+3)}{3}.
  \label{eq:psi_formulation}
  \end{equation}
  Since each kernel satisfies $\nabla\cdot\psi_{\mathrm{iso}}=\mathbf{0}$, any linear combination remains exactly divergence-free.
\begin{equation} 
    \vfld(\vx)=\vmean +\sum_{k=1}^{K}\psi_{\mathrm{iso}}(\vx-\kcenk;\kradk)\,\kwgtk. 
    \label{eq:dfk} 
\end{equation} 
The constant term $\vmean$ captures global transport while preserving incompressibility. The resulting parameterization $\Theta=\{\vmean\}\cup\{(\kcenk,\kradk,\kwgtk)\}_{k=1}^{K}$ defines a continuous divergence-free velocity field that can be evaluated at arbitrary query locations. In this work, DFKs serve as the structured output representation of our model: rather than predicting velocities directly, the network predicts the parameters $\Theta$ of the kernel field. This preserves the exact incompressibility guarantees of the DFK formulation while providing a compact and interpretable representation of the flow.
\section{Method}
\label{sec:method}
Given a sparse, noisy, and unordered set of velocity observations $\gO=\{(\obspm,\obsum, \obscm)\}_{m=1}^{M}$, our goal is to reconstruct a continuous divergence-free velocity field. Instead of fitting a divergence-free kernel field independently for each observation set, we learn a direct mapping from observations to kernel parameters: $\mathcal{F}_{\theta}:\gO\rightarrow\Theta$, where $\theta$ denotes the network parameters and $\Theta$ the DFK representation parameters introduced in Section~\ref{sec:background}. The predicted parameters are then evaluated through Eq.~\ref{eq:dfk}, guaranteeing incompressibility by construction. As illustrated in Fig.~\ref{fig:pipeline}, DFK-Former combines a Perceiver-style transformer with a structured divergence-free kernel decoder. Rather than directly predicting velocities, the network predicts the parameters of a DFK field, yielding a physically interpretable latent representation in which each latent token corresponds to a kernel in the final velocity reconstruction. 

\subsection{DFK-Former Architecture}
\paragraph{Observation Encoding and Kernel Queries.} 
Each observation is embedded independently as $\dfkc{dfkconf}{\vo_m} = \mathrm{MLP} \left( [\gamma(\obspm)\Vert\obsum\Vert \obscm] \right)$, where $\gamma(\cdot)$ denotes a Fourier positional encoding and $\obscm$ is an optional confidence indicator. In addition to observation tokens, the model maintains a set of $K$ kernel-query tokens $\{\dfkc{dfkpos}{\vq_k}\}_{k=1}^{K}$ and a global register token $\dfkc{dfkmean}{\vr}$. Each query token is initialized from a learnable embedding together with the Fourier encoding of an initial kernel center $\kceninit$, establishing an explicit correspondence between latent tokens and physical kernel locations. The transformer backbone alternates between cross-attention and self-attention, as shown in Fig.~\ref{fig:pipeline}. During cross-attention, kernel-query tokens attend to the observation tokens and gather evidence from the sparse measurements. During self-attention, kernel and register tokens exchange information and coordinate their explanations of the observed flow. The resulting kernel-query states therefore summarize both local observations and global flow structure. Architecture hyperparameters are listed in Table~\ref{tab:impl-arch}.\\
\paragraph{Fixed-Kernel DFK Prediction.} We first consider the DFK representation of Eq.~\ref{eq:dfk} with fixed kernel locations and support radii. Kernel centers are initialized using Poisson-disk sampling over the reconstruction domain~\citep{ni2025representing}, while support radii follow a predefined (optionally multi-scale) configuration. The prediction heads decode kernel coefficients and a global mean-flow component from the transformer outputs: $\kwgtk=f_{\omega}(\dfkc{dfkpos}{\vq_k})$ and $\vmean=f_{\mathrm{mean}} (\dfkc{dfkmean}{\vr})$. The reconstructed field $\vfld$ is then obtained from Eq.~\ref{eq:dfk}. This formulation replaces per-instance estimation of DFK coefficients with a learned predictor. For fixed kernel locations and support radii, the field is linear in $\{\kwgtk\}$ and $\vmean$, and these parameters can be recovered by least-squares fitting under dense, clean observations. In the sparse, noisy, and partial-observation regime, however, the inverse problem becomes ill-conditioned. DFK-Former instead learns a data-driven prior that directly predicts the kernel coefficients and mean-flow term from observations in a single forward pass.\\
\paragraph{Adaptive Kernel Placement.} 
\label{sec:adaptive} 
Fixed kernel layouts cannot easily adapt to highly non-uniform flow structures. To increase flexibility, each kernel-query state directly predicts a kernel displacement and support radius, with \(\kcenk=\kceninit+\kcendel\) and \(\kradk=\dfkc{dfkradius}{\sigma_k^{0}}\dfkc{dfkradius}{\rho_k}\).
These quantities are decoded directly from the corresponding kernel-query state. The displacement and radius heads are initialized such that $\kcendel=\mathbf{0}$ and $\dfkc{dfkradius}{\rho_k}=1$, ensuring that the adaptive model initially behaves identically to the fixed-kernel formulation. Because DFKs remain divergence-free for arbitrary kernel locations and support radii, adaptive placement improves reconstruction capacity without sacrificing the physical guarantees of the underlying representation.\\
\paragraph{Anisotropic Divergence-Free Kernels.} 
\label{sec:aniso-method} 
The kernels introduced so far are isotropic and therefore model flow structures using spherical regions of influence. However, many physically relevant structures, such as vortex tubes and shear layers, exhibit strong directional anisotropy. To capture such behavior, we replace isotropic supports with oriented ellipsoidal kernels. Each kernel is parameterized by a rotation matrix $\krotk\in\mathrm{SO}(3)$ and strictly positive physical semi-axis lengths $\kaspk$. Let $\psi_{\mathrm{ref}}(\boldsymbol\xi):=\psi_{\mathrm{iso}}(\boldsymbol\xi;1)$ denote the unit-radius kernel of Eq.~\ref{eq:psi_formulation}. Following the Piola transform, we define the spatial profile of kernel $k$ as follows: 
\begin{equation}\label{eq:aniso-kernel}
    \psi_k(\vz) := \det(\kmap)^{-1} \kmap \psi_{\mathrm{ref}} \!\left( \kmap^{-1}\vz \right), \qquad \kmap = \krotk \operatorname{diag}(\kaspk). 
\end{equation}
Eq.~\ref{eq:aniso-kernel} is the contravariant Piola transformation associated with the linear map $\mathbf z=\kmap\boldsymbol{\xi}$. The Piola transformation commutes with the divergence operator and, therefore, maps divergence-free reference fields to divergence-free physical fields~\citep{boffi2013mixed}. Consequently,
$
\nabla_{\mathbf z}\cdot
\left[\psi_k(\mathbf z)\dfkc{dfkweight}{\boldsymbol{\omega}}\right]=0
$
$
\forall\,\dfkc{dfkweight}{\boldsymbol{\omega}}\in\mathbb{R}^3,
$
for any invertible $\kmap$. A direct derivation is provided in Appendix \ref{app:piola_divergence}.
A key advantage of this construction is that divergence-freeness is preserved for any invertible transformation $\kmap$. Consequently, the network can learn kernel orientation and anisotropy directly from data while retaining exact incompressibility. In practice, we employ an axisymmetric parameterization $\kaspk=(\dfkc{dfkaspect}{a_{\perp,k}},\dfkc{dfkaspect}{a_{\perp,k}},\dfkc{dfkaspect}{a_{\mathrm{long},k}})$. Under our extend-only convention, $\dfkc{dfkaspect}{a_{\perp,k}}=\kradk$ and $\dfkc{dfkaspect}{a_{\mathrm{long},k}}=\kradk\dfkc{dfkaspect}{\eta_k}$, where $\dfkc{dfkaspect}{\eta_k}\ge1$ is the aspect ratio. Thus $\kaspk$ already includes the radius scale: no additional $\kradk$ is applied to $\psi_{\mathrm{ref}}$. For $\krotk=\mathbf I$ and $\kaspk=\kradk(1,1,1)$, $\psi_k(\vz)=\psi_{\mathrm{iso}}(\vz;\kradk)$, including its $\kradk^{-2}$ normalization. The anisotropic field uses $\psi_k(\vx-\kcenk)$ in place of $\psi_{\mathrm{iso}}(\vx-\kcenk;\kradk)$ in Eq.~\ref{eq:dfk}.

\subsection{Temporal DFK-Former}
\label{sec:temporal-method}
We extend DFK-Former to time-varying flows by operating on sequences $\{(\gO_n,t_n)\}_{n=1}^{T}$, where $n$ indexes frames, $t_n\in[0,1]$ is the normalized acquisition time, and $\gO_n=\{(\dfkc{dfkpos}{\vp_{n,m}},\dfkc{dfkvel}{\vu_{n,m}},\dfkc{dfkconf}{c_{n,m}})\}_{m=1}^{M_n}$. The timestamp $t_n$ is shared by all observations in frame $n$; the input tuple in Fig.~\ref{fig:pipeline} records this association. Observation embeddings use the same position, velocity, and confidence inputs as the static model; the timestamp is not an additional input to that observation MLP. We reserve $t\in[0,1]$ for continuous query time. Our key design principle is that kernel identities are preserved across time: the same $K$ kernel-query tokens represent persistent flow structures throughout the sequence, eliminating the need for explicit kernel matching between frames. For each frame, observation tokens are computed as in the static model. Kernel queries attend to the observations through cross-attention and subsequently exchange information through alternating spatial and temporal self-attention. Spatial attention allows kernels within a frame to coordinate their explanations of the flow, while temporal attention propagates information across frames and promotes consistent kernel identities. The resulting per-frame kernel states are mean-pooled over time and decoded into Fourier coefficients describing continuous coefficient trajectories $\kwgtkt$ and the mean-flow term $\vmeant$:
\begin{equation}
    \begin{bmatrix} \kwgtkt\\ \vmeant \end{bmatrix} = \begin{bmatrix} \dfkc{dfkweight}{\alpha_k^{0}}\\ \dfkc{dfkmean}{\bar{\alpha}^{0}} \end{bmatrix} + \sum_{f=1}^{F} \left( \begin{bmatrix} \dfkc{dfkweight}{\alpha_k^{f}}\\ \dfkc{dfkmean}{\bar{\alpha}^{f}} \end{bmatrix} \cos(2\pi f \tvar) + \begin{bmatrix} \dfkc{dfkweight}{\beta_k^{f}}\\ \dfkc{dfkmean}{\bar{\beta}^{f}} \end{bmatrix} \sin(2\pi f \tvar) \right).
\end{equation}
This low-order Fourier parameterization provides a compact continuous-time representation and naturally encourages temporal smoothness.
\subsection{Procedural Training}
\label{sec:data} 

\textbf{Training Data Generation.}
All training examples are generated on the fly rather than drawn from a fixed dataset. For each training step, we sample a clean divergence-free target field $\vstar$ together with a corrupted observation set $\gO$. This procedure prevents memorization of a finite training set and allows continuous variation of flow complexity, sampling density, and corruption severity. Our primary field generator samples Gaussian random fields on a periodic Fourier grid and projects them onto the divergence-free subspace using an exact Helmholtz decomposition in Fourier space, 
\begin{equation}
    \widehat{\vstar}(\vk) = (I-\hat{\vk}\hat{\vk}^{\!\top})\dfkc{dfkvel}{\widehat{\vu}}(\vk), 
\end{equation} 
ensuring incompressibility to numerical precision  (Appendix~\ref{app:data-generation}, Fig.~\ref{fig:data-pipeline}). Corrupted observations are generated by adding a controlled irrotational component and Gaussian noise, then
applying sparse sampling and occlusion masks. The target remains the clean divergence-free field $\vstar$. To complement the spectral generator, we introduce a \emph{Composite Analytic-Flow} (CAF) benchmark, which randomly combines multiple analytic flows, including Lamb--Oseen vortices, Taylor--Green vortices, ABC flows, shear layers, and vortex rings, with randomly sampled mixing weights. These flows introduce coherent structures that are difficult to obtain from Gaussian random fields alone. 

\textbf{Capacity-Matched Sampling.}
Since DFK-Former predicts a continuous field representation, supervision is performed at randomly sampled spatial query points rather than on a fixed grid. During training, we randomize the number of observations, sampling patterns, noise levels, and flow spectra to expose the model to a wide range of reconstruction conditions. To avoid introducing irreducible approximation error, the spatial spectrum of the generated fields is band-limited according to the representational capacity of the kernel basis (Appendix~\ref{app:data-capacity}). This ensures that training targets remain expressible by the chosen kernel configuration, allowing optimization to focus on inference from sparse observations rather than compensating for unattainable high-frequency content.

\textbf{Training Objective.} 
The primary supervision matches the reconstructed field at query locations, $\gL_{\mathrm{field}}=\mathbb{E}_{\vx\sim Q}\!\left[\|\vfld(\vx)-\vstar(\vx)\|_2^2\right]$. In addition, we enforce consistency in the observed samples $\gL_{\mathrm{data}}=\frac{1}{M}\sum_{m=1}^{M}\obscm\|\vfld(\obspm)-\vstar(\obspm)\|_2^2$. Both terms are Huberized once strong corruption is introduced to reduce sensitivity to outliers. Two field-level regularizers further improve reconstruction quality. A magnitude-weighted cosine loss $\gL_{\mathrm{cos}}$ improves directional accuracy, particularly under sparse observations, while an orthogonality loss $\gL_{\mathrm{ortho}}$ discourages alignment with the known irrotational contamination at observed samples. The cosine term is introduced gradually, and $\gL_{\mathrm{ortho}}$ is activated only under severe contamination. Additional losses regularize the geometry heads when enabled. For adaptive placement, $\gL_{\mathrm{ctr}}$ and $\gL_{\mathrm{rad}}$ keep predicted centers and radii close to their anchors, while a coverage term $\gL_{\mathrm{cov}}$ prevents kernel collapse by softly repelling centers. For anisotropic kernels, a warm-start loss $\gL_{\mathrm{sh}}$ initializes orientations and aspect ratios from sample-based estimates and is annealed to zero early in training. The final objective is a weighted sum of all these losses, with stage-specific weights provided in Table~\ref{tab:impl-train} and  Table~\ref{tab:data-stages}. Training follows a corruption curriculum, progressing from lightly corrupted observations to heavy noise, sparse sampling, and large occlusions.

\textbf{Temporal Training.} 
The temporal model is trained using procedurally generated time-varying divergence-free flows with two complementary generators. The first form trajectories by combining divergence-free spatial modes with time-varying Fourier coefficients, $\vv(\vx,\tvar)=\sum_j c_j(\tvar)\,\vv_j(\vx)$, while the second models transported coherent structures through collections of advected divergence-free primitives (Appendix~\ref{app:data-temporal}). Both preserve incompressibility for all $\tvar$. Observation counts, noise realizations, and occlusion masks are sampled independently for each frame. We supervise in randomly sampled space-time query locations using $\gL_{\mathrm{field}}=\mathbb{E}_{(\vx,\tvar)}\left[\|\vfld(\vx,\tvar)-\vstar(\vx,\tvar)\|_2^2\right]$, with the corresponding observation-consistency loss evaluated in the sampled measurements. Because Temporal DFK-Former represents kernel coefficients using a Fourier expansion of order $F$ (Sec. \ref{sec:temporal-method}), the training data's temporal spectrum is band-limited to match model capacity. This avoids supervising temporal frequencies that the chosen Fourier basis cannot represent and enables a controlled study of the trade-off between temporal bandwidth and reconstruction accuracy.
\section{Results} 
\label{sec:results} 

\subsection{Experimental Setup} 
\label{sec:exp-setup} 
\textbf{Datasets.} We evaluate DFK-Former on $582$ synthetic and real incompressible flow fields, drawn from four groups: (1) Gaussian Random fields, (2) the CAF set from Section~\ref{sec:data} (drawn from the training distribution), (3) five analytic families with closed-form ground truth (Taylor--Green, Lamb--Oseen, vortex ring, shear layer, ABC), and (4) four real-DNS JHTDB sets~\citep{li2008public} (Isotropic turbulence, Forced MHD turbulence, Channel Flow, and Transitional Boundary Layer). Inputs are corrupted with varying sparsity ($M\!\in\![187,4086]$), noise ($\mathrm{RMS}\!\in\![0,15\%]$), irrotational contamination ($\chi\!\in\![0,1]$), and occlusion (0--35\%), while targets remain clean Leray-projected velocity fields. For temporal reconstruction, we evaluated on real DNS sequences, comprising 60 time-resolved JHTDB flows with three unseen midpoint frames per sequence.\\
\textbf{Metrics.} Performance is evaluated through reconstruction accuracy (EPE, relative $L^2$, MAE, cosine similarity, angular error), physical consistency (divergence), and computational efficiency (latency, throughput, memory, FLOPs, and parameter count).\\
\textbf{Baselines.} We compare DFK-Former against optimization-based methods (Wendland RBF scalar interpolation, SIREN/Curl-SIREN~\citep{sitzmann2020implicit}, vector-potential INRs~\citep{richterpowell2022neural}, and DFK fitting ~\citep{ni2025representing}), operator-learning approaches (FNO and DeepONet), and physics-constrained models (PINNs with soft divergence penalties, and HDNet-3D~\footnote{We use a 3D version of HDNet~\citep{qi2024hdnet}, trained on the same dataset as DFK-Former.}). We report two DFK variants: \emph{fixed-kernels}, which fits only kernel weights on a predefined kernel layout, and \emph{adaptive-kernels}, which jointly optimizes kernel positions, radii, and weights for each field. For the temporal extension, we compare against Space-Time (ST-) SIREN/Curl-SIREN, per-frame DFK fitting, and the shared-kernel sequence fitting approach proposed by~\citet{ni2025representing}.  Implementation details are given in Appendix~\ref{app:impl}.

\subsection{Main Benchmark} 
\begin{table}[hbp]
\caption{Main reconstruction benchmark. Reconstruction accuracy, divergence residual, and per-field inference cost of DFK-Former against optimization-based, operator-learning, and physics-constrained baselines. Best in \textbf{bold}, second-best \underline{underlined}.}
\label{tab:exp-main}
\centering
\resizebox{\textwidth}{!}{
\begin{tabular}{lcccccccc}
\toprule
Method & RMSE/\ensuremath{\EPE} $\downarrow$ & rel.\ $L^2$ $\downarrow$ & MAE $\downarrow$ & cosine $\uparrow$ & angle ($^\circ$) $\downarrow$ & $\lVert\nabla\!\cdot\!v\rVert_2$ $\downarrow$ & ms/field $\downarrow$ & div-free? \\
\midrule
Wendland RBF (scalar)                      & 0.598 & 0.736 & 0.280 & 0.663 & 37.9 & $3.2$ & 26 & no \\
SIREN ($+$ soft penalty)                       & 0.392 & 0.519 & 0.195 & 0.805 & 26.7 & $1.4$ & 53600 & no (soft) \\
Curl-SIREN                                 & 0.950 & 1.352 & 0.475 & 0.629 & 42.0 & $\sim\!10^{-14}$ & 52700 & by construction \\
Vector-potential INR                       & 0.778 & 1.092 & 0.388 & 0.661 & 39.1 & $\sim\!10^{-15}$ & 48000 & by construction \\
DFK fitting (fixed kernels)             & 0.689 & 0.924 & 0.335 & 0.482 & 52.9 & $\sim\!10^{-15}$ & 22 & by construction \\
DFK fitting (adaptive kernels)          & 0.546 & 0.719 & 0.264 & 0.568 & 45.5 & $\sim\!10^{-15}$ & 9821 & by construction \\
FNO                                        & 0.244 & 0.312 & 0.121 & 0.900 & 16.9 & $0.71$ & 4.3 & no \\
FNO ($+$ Leray projection)                 & 0.251 & 0.318 & 0.124 & 0.896 & 17.4 & $\sim\!10^{-16}$ & 4.9 & by construction \\
DeepONet                                   & 0.417 & 0.509 & 0.203 & 0.764 & 30.5 & $0.23$ & \textbf{0.5} & no \\
PINN ($+$ soft penalty)                        & 0.319 & 0.422 & 0.157 & 0.838 & 22.6 & $0.44$ & 51000 & no (soft) \\
HDNet-3D                                   & \underline{0.211} & \underline{0.257} & \underline{0.105} & \underline{0.935} & \underline{13.7} & $0.66$ & \underline{4.0} & partially \\
\textbf{DFK-Former (ours)}                 & \textbf{0.120} & \textbf{0.151} & \textbf{0.059} & \textbf{0.964} & \textbf{8.0} & $\sim\!10^{-7}$ & 7.5 & by construction \\
\bottomrule
\end{tabular}}
\end{table}
Table~\ref{tab:exp-main} compares all methods on the 582-flow benchmark in terms of reconstruction accuracy, divergence preservation, and inference cost. Optimization-based DFK fitting highlights the accuracy-computation trade-off: fixed-kernel fitting\footnote{Equivalent to an $L^2$ least-squares fit over a fixed kernel basis.} is constrained by a predefined basis, whereas adaptive-kernel fitting improves accuracy through per-field optimization at substantially higher cost. Exactly divergence-free INRs such as Curl-SIREN and vector-potential models preserve incompressibility but remain challenging to fit from sparse and corrupted observations. Operator-learning baselines achieve stronger reconstruction accuracy, with HDNet-3D emerging as the strongest prior method and FNO offering a favorable accuracy-speed trade-off. Applying a Leray projection to FNO (and similarly to DeepONet) eliminates divergence but does not improve reconstruction accuracy. Thus, enforcing incompressibility after prediction cannot recover errors already present in the velocity field. While HDNet-3D projects reconstructed flows toward a divergence-free space, discretization on a voxel grid still introduces residual numerical divergence. DFK-Former achieves the best performance across all reconstruction metrics. Despite enforcing incompressibility to numerical precision ($\sim10^{-7}$), it requires only 7.5 ms per field, comparable to fast feed-forward operator-learning models and roughly three orders of magnitude faster than optimization-based divergence-free approaches. These results demonstrate that directly predicting adaptive DFK parameters combines the accuracy and physical guarantees of divergence-free representations with the efficiency of feed-forward inference.

\subsection{Time-varying flow}
\label{sec:temporal}
\begin{figure}[!t]
\centering
\includegraphics[width=0.85\linewidth]{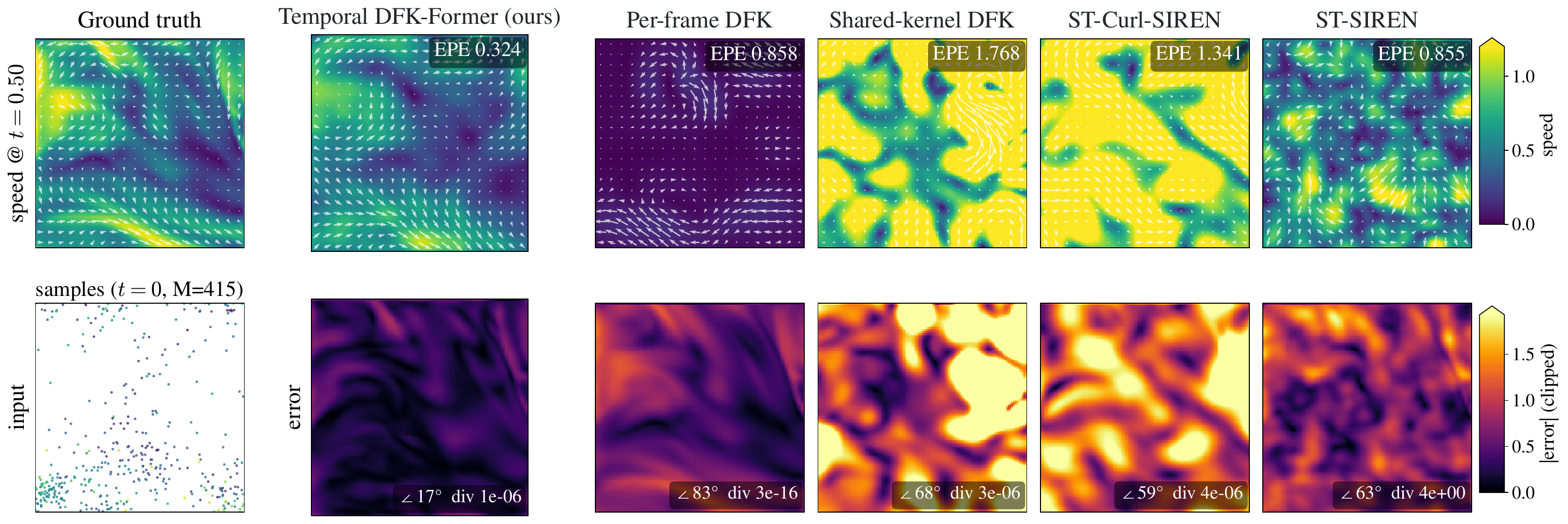}
\caption{\textbf{Temporal reconstruction in the hardest evaluation regime} (Forced MHD (dynamic JHTDB), \texttt{hard combo}: sparse observations, noise, occlusions, and outliers). Top: ground truth and predictions from Temporal DFK-Former, per-frame DFK fitting, iterative DFK fitting, space-time Curl-SIREN, and space-time SIREN. Bottom: input observations and reconstruction errors. DFK-Former remains robust under severe degradation, recovering the dominant flow structures with substantially lower EPE and angular error than competing methods.}
\label{fig:temporal-baselines-hardcombo}
\end{figure}

Table~\ref{tab:temporal-jhtdb} shows that the temporal extension of DFK-Former achieves the lowest overall median EPE (0.114), outperforming per-frame DFK fitting (0.767), shared-kernel DFK fitting (0.626), and Space-Time SIREN/Curl-SIREN (0.428/0.842), while preserving incompressibility to numerical precision. It also achieves the lowest mean EPE (0.124) and mean angular error ($4.8^\circ$). Gains are particularly pronounced under sparse and occluded observations, where temporal coherence substantially improves reconstruction quality, reducing error by up to a factor of $9.7$ compared to per-frame fitting. Figure~\ref{fig:temporal-baselines-hardcombo} illustrates the most challenging regime with combined sparse, occluded, and corrupted settings. Notably, the mean error of per-frame DFK fitting is heavily affected by a small number of near-singular optimization failures. Additional regimes are reported in Appendix~\ref{app:res-temporal}.
\vspace{-0.45cm}
\begin{table}[ht!]
\caption{Temporal flow reconstruction on 60 time-resolved JHTDB flows. Median \ensuremath{\EPE} by observation regime, plus overall mean \ensuremath{\EPE} and angular error; lower is better. Best results are in \textbf{bold}, second-best \underline{underlined}.}
\label{tab:temporal-jhtdb}
\centering
\resizebox{\textwidth}{!}{
\begin{tabular}{lccccc}
\toprule
Regime & \textbf{Temporal DFK-Former (ours)} & Per-frame DFK & Shared-kernel DFK & ST-SIREN & ST-Curl-SIREN \\
\midrule
\texttt{clean\&dense} & 0.114 & \underline{0.100} & \textbf{0.083} & 0.442 & 0.781 \\
\texttt{sparse}           & \textbf{0.109} & 0.847 & 1.093 & \underline{0.388} & 0.590 \\
\texttt{occluded}         & \textbf{0.098} & 0.951 & 1.511 & \underline{0.413} & 0.766 \\
\midrule
Overall (median \ensuremath{\EPE}) & \textbf{0.114} & 0.767 & 0.626 & 0.428 & 0.842 \\
Overall (mean \ensuremath{\EPE})   & \textbf{0.124} & 20.29$^\dagger$ & 0.762 & 0.466 & 0.821 \\
Overall (mean angular error)               & \textbf{4.8$^\circ$} & 40.0$^\circ$ & 25.4$^\circ$ & 21.3$^\circ$ & 37.1$^\circ$ \\
\bottomrule
\end{tabular}}
\\[3pt]
\noindent
\begin{minipage}{\textwidth}
\raggedright\scriptsize $^\dagger$Per-frame DFK's mean is inflated by $15/180$ near-singular per-frame solves.
\end{minipage}
\end{table}

\subsection{Ablations} 
\label{sec:ablations}

\textbf{Representation vs. Backbone.} To isolate the role of the DFK representation, we replace the DFK parameter head with a decoder that directly predicts velocity values while keeping the backbone and training protocol unchanged. The direct regressor achieves similar accuracy to DFK-Former (EPE $0.127$ vs.\ $0.120$) at comparable latency ($6.8$ vs.\ $7.5$ ms), but produces substantial divergence ($\|\nabla\!\cdot\!v\|_2 = 0.26$). This indicates that reconstruction accuracy primarily stems from the learned prior, while the DFK output space enforces incompressibility without sacrificing accuracy. Projecting the regressed field into the same DFK basis restores exact divergence-free structure but increases error (EPE $0.165$) and runtime ($29~$ms), highlighting the benefit of directly predicting DFK parameters.\\
\textbf{The Role of Attention.} Replacing the attention-based encoder with permutation-invariant DeepSets~\citep{zaheer2017deep} pooling while keeping all other components fixed significantly degrades reconstruction quality (EPE $0.558\!\rightarrow\!0.832$, angular error $42^\circ\!\rightarrow\!76^\circ$). Although evaluated in a reduced-scale setting, this result suggests that modeling interactions between observations is essential, whereas simple symmetric pooling is insufficient despite preserving permutation invariance.\\
\textbf{Representation Ladder Analysis.} Table~\ref{tab:ladder} presents cumulative ablations of the kernel parameterization. Adaptive placement yields modest gains, whereas multi-scale radii provide the largest benefit, reducing CAF EPE from $0.171$ to $0.102$ and improving performance across all datasets.\\
\textbf{Kernel Count.} Table~\ref{tab:exp-ablation-K} varies the number of kernels $K$ while keeping all other settings fixed. Accuracy improves from $K=32$ to $1024$ (EPE $0.281\!\rightarrow\!0.120$), then degrades for larger budgets ($0.144$ at $K=2048$). An independent $K=8192$ run achieves a similar error, suggesting limited benefit from further increasing basis resolution. Inference cost remains nearly constant across a $32\times$ range of $K$ since only the query tokens scale with the kernel count. Notably, even the smallest model ($K=32$) outperforms closed-form DFK fitting, indicating that the gains arise primarily from the learned prior rather than basis size alone.\\
\textbf{Anisotropic Kernels.} To isolate representational gains from parameter-count effects, we compare the anisotropic model ($K=1024$) with a parameter-matched isotropic variant ($K=1542$). Table~\ref{tab:exp-ablation-aniso} shows consistent improvements across all analytic flows, with gains increasing from $1.16\times$ on Lamb--Oseen vortices to $1.34\times$ on vortex rings.
\vspace{-0.4cm}

\begin{table*}[ht!]
\centering
\scriptsize
\setlength{\tabcolsep}{2.5pt}
\renewcommand{\arraystretch}{0.63}
\begin{minipage}[c]{0.52\linewidth}
\centering

\caption{\textbf{Component ladder.} Progressively adding adaptive placement and multi-scale radii to the anisotropic base model. Multi-scale radii provide the largest gains, particularly on CAF flows and real DNS data. Lower is better.}
\label{tab:ladder}

\vspace{0.2em}

\resizebox{\linewidth}{!}{
\begin{tabular}{l cc@{\hskip 0.9em} cc@{\hskip 0.9em} cc}
\toprule
& \multicolumn{2}{c}{in-dist} & \multicolumn{2}{c}{CAF} & \multicolumn{2}{c}{JHTDB (real DNS)} \\
\cmidrule(lr){2-3}\cmidrule(lr){4-5}\cmidrule(lr){6-7}
Configuration & \ensuremath{\EPE}$\downarrow$ & ang.$^\circ\!\downarrow$
              & \ensuremath{\EPE}$\downarrow$ & ang.$^\circ\!\downarrow$
              & \ensuremath{\EPE}$\downarrow$ & ang.$^\circ\!\downarrow$ \\
\midrule
Anisotropic (base)         & $0.382$ & $28.3$ & $0.164$ & $11.4$ & $0.167$ & $9.3$ \\
\ $+$\,adaptive placement  & $0.368$ & $27.4$ & $0.171$ & $12.2$ & $0.162$ & $9.1$ \\
\ \ $+$\,multi-scale radii & $\mathbf{0.343}$ & $\mathbf{25.5}$ & $\mathbf{0.102}$ & $\mathbf{6.9}$ & $\mathbf{0.155}$ & $\mathbf{8.7}$ \\
\bottomrule
\end{tabular}
}

\end{minipage}
\hfill
\begin{minipage}[c]{0.45\linewidth}
\centering

\caption{\textbf{Kernel-Count Ablation.} Scaling the number of kernels improves reconstruction quality up to $K{=}1024$, which achieves the best accuracy-latency trade-off. Lower is better.}
\label{tab:exp-ablation-K}

\vspace{0.2em}

\resizebox{\linewidth}{!}{
\begin{tabular}{lccccccc}
\toprule
$K$ & $32$ & $64$ & $128$ & $256$ & $512$ & $1024$ & $2048$ \\
\midrule
\ensuremath{\EPE}$\downarrow$ & $0.281$ & $0.226$ & $0.200$ & $0.193$ & $0.142$ & $\mathbf{0.120}$ & $0.144$ \\
angle ($^\circ$)$\downarrow$ & $20.8$ & $16.1$ & $14.0$ & $13.4$ & $9.6$ & $\mathbf{8.0}$ & $9.7$ \\
ms/field$\downarrow$         & $\mathbf{5.7}$ & $\mathbf{5.7}$ & $5.9$ & $5.9$ & $6.0$ & $7.5$ & $11.6$ \\
peak mem (MB)$\downarrow$    & $\mathbf{58}$ & $\mathbf{58}$ & $59$ & $59$ & $72$ & $72$ & $80$ \\
\bottomrule
\end{tabular}
}

\end{minipage}
\end{table*}
\vspace{-0.4cm}
\begin{table}[!ht]
\caption{\textbf{Anisotropy vs.\ a parameter-matched isotropic control} on
structured analytic flows: our anisotropic $K{=}1024$ model against a
separately trained isotropic model at $K{=}1542$, matched to within four
scalar parameters. rel\ensuremath{\EPE}, lower is better; improvement is the
isotropic/anisotropic ratio.}
\label{tab:exp-ablation-aniso}
\centering
\small
\begin{tabular}{lccc}
\toprule
Target structure & Isotropic & Anisotropic & improvement \\
\midrule
Vortex tube (Lamb--Oseen) & 0.0763 & \textbf{0.0656} & $1.16\times$ \\
Shear layer               & 0.0719 & \textbf{0.0592} & $1.22\times$ \\
Vortex ring               & 0.0835 & \textbf{0.0623} & $1.34\times$ \\
\bottomrule
\end{tabular}
\end{table}

\section{Conclusion and Limitations} 
\label{sec:conclusion} 
We presented \textbf{DFK-Former}, a set-to-field transformer that predicts the parameters of a divergence-free kernel representation from sparse, noisy, and irregular observations. The resulting velocity fields are exactly divergence-free, mesh-free, and continuously queryable by construction, while avoiding per-instance optimization at inference time. Experiments across diverse reconstruction settings demonstrate that direct prediction of divergence-free kernel parameters combines the accuracy and physical guarantees of kernel-based representations with the efficiency and robustness of learned flow priors. Anisotropic kernels and temporal modeling further extend the framework to more complex spatial and temporal flow structures while preserving exact incompressibility. 

Limitations include the use of synthetic training data and the limited exploration of larger model and kernel capacities. Although the learned representation generalizes to DNS turbulence without retraining, broader evaluation on experimental measurements and larger-scale studies are needed to better characterize robustness and scaling behavior.

\subsubsection*{AI use statement}
In this work, we used large language models (LLMs) for four main tasks, all under author supervision.

\textbf{Retrieval and discovery:} to suggest candidate related work; every resulting reference was then located, read, and verified against the primary source by the authors, and any suggestion that could not be verified was discarded.

\textbf{Research ideation and execution:} to brainstorm design alternatives, and as a coding assistant for implementing and debugging the model, baselines, data generators, and evaluation process, as well as for summarizing the resulting experiment logs.

\textbf{Drafting:} to produce first drafts of expository passages, which the authors then restructured, rewrote, and fact-checked against the underlying results.

\textbf{Writing:} to polish the grammar, phrasing, and clarity of the paper text, as well as to shorten some paragraphs when needed.

The feed-forward divergence-free kernel formulation and overall research direction are the authors' contributions. The outputs of LLMs were reviewed, corrected, and approved by the authors. All reported results were generated by executing the codebase and are supported by experiment logs and checkpoints; no numerical result, table, or figure was generated or estimated by an LLM. The codebase is covered by automated tests, and the authors take full responsibility for the content of this paper.

\subsubsection*{Reproducibility statement}
\textbf{Code, Training and Evaluation Datasets.}
Upon acceptance, we will release the complete codebase, including the DFK kernel layer, DFK-Former, training and evaluation pipelines, procedural data generators, configurations, and scripts used to reproduce all reported results. Training data are generated on the fly from released procedural generators and corruption pipelines using fully specified configurations and random seeds. Synthetic evaluation sets are deterministically generated and distributed with accompanying metadata to ensure identical benchmarking across methods. For real-data experiments, we provide scripts and manifests that reproduce the exact JHTDB~\citep{li2008public} evaluation protocol for users with valid database access. Since JHTDB is a third-party resource, we do not redistribute the data; access requires an individual authorization token issued by the database. The released scripts and metadata enable users with valid credentials to reconstruct the exact evaluation datasets and preprocessing pipeline used in our experiments.

\textbf{Environment.} All experiments were run with PyTorch on NVIDIA A100 GPUs; the released
repository pins the library versions and provides the environment specification, the checkpoints
required to reproduce the reported evaluations, and the commands used to launch each run.

\subsubsection*{Acknowledgments}
For computer time, this research used Ibex managed by the Supercomputing Core Laboratory at King Abdullah University of Science \& Technology (KAUST) in Thuwal, Saudi Arabia

\bibliography{references}
\bibliographystyle{iclr2027_conference}

\clearpage
\appendix
\begin{center}
{\Large\bfseries Appendix: Supplementary Material}\\[0.4em]
\end{center}
\vspace{0.5em}

\section{Divergence Preservation under the Piola Transformation}
\label{app:piola_divergence}

The anisotropic kernel of Eq.~\ref{eq:aniso-kernel} is a contravariant Piola transform of the isotropic kernel, the standard divergence-preserving map used to build $H(\mathrm{div})$-conforming finite elements~\citep{boffi2013mixed,rognes2009efficient}. We restate the identity and verify it directly for our construction.

\paragraph{Proposition.} Let $\psi_{\mathrm{ref}}(\boldsymbol\xi):=\psi_{\mathrm{iso}}(\boldsymbol\xi;1)\in\mathbb{R}^{3\times3}$ be the unit-radius kernel of Eq.~\ref{eq:psi_formulation}, which is divergence-free in every column, and fix a kernel index $k$. Let $\kmap\in\mathbb{R}^{3\times3}$ with $\det(\kmap)>0$ be invertible. Define
\begin{equation}
    \psi_k(\vz) = \frac{1}{\det(\kmap)}\,\kmap\,\psi_{\mathrm{ref}}\!\left(\kmap^{-1}\vz\right).
    \label{eq:piola_kernel_appendix}
\end{equation}
Then $\psi_k$ is divergence-free in every column: $\nabla_{\vz}\cdot[\psi_k(\vz)\kwgt]=0$ for all $\kwgt\in\mathbb{R}^3$.

\paragraph{Proof.} Write $\boldsymbol\xi=\kmap^{-1}\vz$ and, for an arbitrary reference field $\widehat{\vv}$, let $\vv(\vz)=\det(\kmap)^{-1}\kmap\widehat{\vv}(\boldsymbol\xi)$. Since $\kmap$ is constant in $\vz$, the chain rule gives, in index notation with implicit summation,
\begin{equation}
    \begin{aligned}
    \nabla_{\vz}\cdot\vv
    &=\frac{\partial v_i}{\partial z_i}
    = \frac{1}{\det(\kmap)}\,(\kmap)_{ij}\,\frac{\partial\widehat v_j}{\partial\xi_\ell}\,(\kmap^{-1})_{\ell i}
    \\
    &= \frac{1}{\det(\kmap)}\,\delta_{\ell j}\,\frac{\partial\widehat v_j}{\partial\xi_\ell}
    = \frac{1}{\det(\kmap)}\,\nabla_{\boldsymbol\xi}\cdot\widehat{\vv},
    \end{aligned}
    \label{eq:piola_div_identity}
\end{equation}
using $(\kmap^{-1})_{\ell i}(\kmap)_{ij}=\delta_{\ell j}$. Taking $\widehat{\vv}(\boldsymbol\xi)=\psi_{\mathrm{ref}}(\boldsymbol\xi)\kwgt$ -- divergence-free by construction of $\psi_{\mathrm{ref}}$ -- Eq.~\ref{eq:piola_div_identity} gives $\nabla_{\vz}\cdot[\psi_k(\vz)\kwgt]=\det(\kmap)^{-1}\nabla_{\boldsymbol\xi}\cdot[\psi_{\mathrm{ref}}(\boldsymbol\xi)\kwgt]=0$ for every $\kwgt$, so every column of $\psi_k$ is divergence-free. \hfill$\blacksquare$

\paragraph{Application to the anisotropic DFK.} With $\kmap=\krotk\operatorname{diag}(\kaspk)$ as in Eq.~\ref{eq:aniso-kernel}, the proposition applies to every kernel $\psi_k$, so any superposition
\begin{equation}
    \vfld(\vx) = \vmean + \sum_{k=1}^{K} \psi_k(\vx-\kcenk)\,\kwgtk
\end{equation}
satisfies $\nabla_{\vx}\cdot\vfld(\vx)=0$ \emph{exactly}, for arbitrary kernel weights, centers, rotations, and strictly positive semi-axis lengths $\kaspk$. Because the map defined by $\kmap$ is linear, the same map also sends the reference kernel's unit-ball support to the ellipsoid $\{\vz:\|\kmap^{-1}\vz\|\le1\}$: anisotropy reshapes the kernel's support without breaking compactness or exact divergence-freedom.
\newpage
\section{Experimental Details}
\label{app:impl}
We provide the full implementation details used in all experiments. Table~\ref{tab:impl-arch} summarizes the DFK-Former architecture for both the static and temporal models, while Table~\ref{tab:impl-train} reports the corresponding training recipe, optimization settings, losses, and hardware configuration.

\begin{table}[!ht]
\caption{DFK-Former architecture, as resolved for the reported static run. The upper block is the
static model; the lower block gives the temporal model's additions and the two places where it
departs from the static geometry. $K$ is the kernel count and $\Delta$ the mean kernel spacing.}
\label{tab:impl-arch}
\centering
\footnotesize
\begin{tabular}{@{}lp{0.62\linewidth}@{}}
\toprule
Component & Setting \\
\midrule
\multicolumn{2}{@{}l}{\emph{Static model: backbone, heads and kernel geometry}} \\
Token width $d_{\mathrm{model}}$ & $256$ \\
Backbone blocks & $6$; each block $=$ one masked cross-attention sublayer $+$ one self-attention
  sublayer \\
Attention heads / head dimension & $8$ / $32$ \\
Feed-forward hidden width & $4\,d_{\mathrm{model}}=1024$, GELU \\
Dropout & $0$ \\
Normalisation & pre-LayerNorm residual blocks; one final LayerNorm on the query set \\
Fourier position encoding $\gamma$ & $6$ bands at frequencies $2^{0}\dots2^{5}$, raw coordinate
  concatenated ($39$ features) \\
Observation tokenizer & $\mathrm{Linear}(43\!\to\!256)\to\mathrm{GELU}\to\mathrm{Linear}(256\!\to\!256)$ \\
Tokenizer input dimension & $39$ (encoded position) $+\,3$ (velocity) $+\,1$ (confidence) $=43$; $42$
  with the confidence channel disabled \\
Kernel-query tokens & $K$ learnable $256$-vectors plus a linear projection of $\gamma(\kceninit)$ \\
Register tokens & $1$ \\
Prediction heads & $\mathrm{Linear}(256\!\to\!256)\to\mathrm{GELU}\to\mathrm{Linear}(256\!\to\!\cdot)$,
  all zero-initialised (weight and bias) \\
Head outputs & $\kwgtk$ ($3$); $\vmean$ ($3$, from the register state); centre ($3$, $\tanh$); radius
  ($1$, $\tanh$); aspect ($1$, $\tanh$); orientation ($3$, $\tanh$) \\
Kernel count $K$ & $1024$ \\
Kernel centres & Poisson-disk (Mitchell best-candidate), seed $12345$, fixed buffer \\
Kernel spacing & $\Delta = \text{box}/K^{1/3}$ on the canonical box of side $2$ \\
Radius mixture & scales $\{5.0,4.0,3.0\}\,\Delta$ at fractions $\{0.3,0.5,0.2\}$ of $K$ (mean
  $\approx 4.1\,\Delta$) \\
Centre displacement bound & $\lVert\kcendel\rVert \le 1.0\,\Delta$ \\
Radius bound & $\kradk\in[\kradfix/3,\;3\kradfix]$ \\
Aspect bound & $a_{\mathrm{long}}/a_\perp\in[\tfrac14,4]$, extend-only ($a_\perp=\kradk$) \\
Anisotropy construction & contravariant Piola transform (Eq.~\ref{eq:aniso-kernel}) \\
Parameters & $10{,}228{,}750$ \\
\midrule
\multicolumn{2}{@{}l}{\emph{Temporal model (Section~\ref{sec:temporal-method})}} \\
Backbone blocks & $6$; each block $=$ spatial cross-attention $+$ spatial self-attention $+$ temporal self-attention \\
Time encoding & $[\tvar,\ \sin(2\pi f\tvar),\ \cos(2\pi f\tvar)]_{f=1}^{4}\to\mathrm{Linear}(9\!\to\!256)$, added to observation and query tokens \\
Temporal aggregation & per-frame kernel states mean-pooled over $T$ before the Fourier heads \\
Fourier order $F$ & $2$ (reported model); $7$ for the ablation arm of Section~\ref{app:abl-F} \\
Frames per example (training) & $T\sim\gU\{6,\dots,12\}$ \\
Kernel geometry & one anisotropic kernel set shared across all frames: a single radius scale
  $3.5\,\Delta$ in place of the static model's mixture, fixed centres and radii, centre advection
  disabled. Only the Fourier weights and $\vmean$ vary in $\tvar$ \\
Parameters & $14{,}843{,}426$; $14{,}976{,}038$ with adaptive placement \\
\bottomrule
\end{tabular}
\end{table}

\begin{table}[!ht]
\caption{Training recipe of the reported static model, read from its fully resolved configuration.
Ablation arms differ from it only in the single factor each ablation names
(Section~\ref{app:ablations}); the temporal model's departures are in the last block. The auxiliary
block acts on the placement and shape heads during training only and has no effect at inference.
Per-stage data distributions are in Table~\ref{tab:data-stages}; supervision point counts are in
Table~\ref{tab:data-params}.}
\label{tab:impl-train}
\centering
\footnotesize
\begin{tabular}{@{}lp{0.64\linewidth}@{}}
\toprule
Setting & Value \\
\midrule
\multicolumn{2}{@{}l}{\emph{Optimization and schedule}} \\
Optimizer & AdamW, library-default $\beta$ and $\epsilon$; weight decay $0$ \\
Peak learning rate & $3\times10^{-4}$ \\
Schedule & constant to step $60{,}000$, cosine decay to step $145{,}000$, held at $10^{-5}$
  thereafter; \textbf{no warm-up} \\
Gradient clipping & global norm $1.0$ \\
Batch size & $32$ fields per rank $\times\ 4$ ranks $=128$ fields per step \\
Steps & $150{,}000$ $=$ Core $60{,}000$ $+$ Hard $85{,}000$ $+$ cooldown $5{,}000$ \\
Precision & model float32; bf16 autocast over the encoder forward only --- kernel evaluation and all
  losses run in float32 \\
Weight averaging & EMA, decay $0.9995$; all reported evaluations use the EMA weights \\
Checkpoint selection & minimum of $\tfrac12\bigl(\EPE_{\text{GRF}}+\EPE_{\text{CAF}}\bigr)$ on
  the frozen validation sets, scored on EMA weights, every $500$ steps \\
\midrule
\multicolumn{2}{@{}l}{\emph{Objective}} \\
Objective & $\gL=\gL_{\mathrm{field}}+\lambda_{\mathrm{data}}\gL_{\mathrm{data}}
  +\lambda_{\mathrm{cos}}\gL_{\mathrm{cos}}+\lambda_{\mathrm{ortho}}\gL_{\mathrm{ortho}}
  +\lambda_{\mathrm{cov}}\gL_{\mathrm{cov}}+\lambda_{\mathrm{ctr}}\gL_{\mathrm{ctr}}
  +\lambda_{\mathrm{rad}}\gL_{\mathrm{rad}}+\lambda_{\mathrm{sh}}\gL_{\mathrm{sh}}$; the last four
  are the auxiliary block below \\
$\lambda_{\mathrm{field}}$ / $\lambda_{\mathrm{data}}$ & $1.0$ / $1.0$ (all stages) \\
$\lambda_{\mathrm{cos}}$ & $0.5$ (all stages), linearly ramped over the first $3{,}000$ steps;
  magnitude floor $0.05$ \\
$\lambda_{\mathrm{ortho}}$ & $0$ (Core), $0.1$ (Hard), $0.1$ (cooldown) \\
Huber $\delta$ on the data term & $1.0$ (Hard and cooldown only) \\
Divergence & never in the objective; diagnostic only \\
\midrule
\multicolumn{2}{@{}l}{\emph{Auxiliary regularisers on the placement and shape heads (training only)}} \\
$\lambda_{\mathrm{ctr}}$ / $\lambda_{\mathrm{rad}}$ & $0.01$ / $0.01$, all stages; keeps the
  predicted centres and radii near their fixed anchors \\
$\lambda_{\mathrm{cov}}$ & $10^{-3}$ on Hard and cooldown; ramped over the first $20{,}000$ steps
  \emph{of the Hard stage}, unramped thereafter. Soft repulsion against centre collapse \\
$\lambda_{\mathrm{sh}}$ (orientation warm-start) & $0.1$ annealed linearly to zero over $5{,}000$
  steps, aspect factor $3.0$; Core only. Target is computed from the input samples, never the
  clean field \\
\midrule
\multicolumn{2}{@{}l}{\emph{Temporal model: departures from the above}} \\
Batch size & $16$ fields per rank $\times\ 4$ ranks $=64$ fields per step \\
Objective & no auxiliary block: placement is fixed, so the coverage, centre, radius and shape terms
  are absent; $\gL_{\mathrm{field}}$, $\gL_{\mathrm{data}}$, $\gL_{\mathrm{cos}}$ and
  $\gL_{\mathrm{ortho}}$ carry the same weights and stage schedule \\
Checkpoint selection & validation \ensuremath{\EPE} alone (no CAF term), EMA weights \\
\midrule
\multicolumn{2}{@{}l}{\emph{Run provenance}} \\
Seed & $0$; data seeds are rank-distinct \\
Hardware & $4\times$ NVIDIA A100 80GB PCIe \\
Wall clock & $28$\,h\,$02$\,m (Core) $+$ $31$\,h\,$20$\,m (Hard) $+$ $2$\,h\,$20$\,m (cooldown), one
  uninterrupted launch \\
\bottomrule
\end{tabular}
\end{table}
\newpage
\section{Data Generation}
\label{app:data-generation}

\begin{figure}[ht!]
\centering
\includegraphics[width=\linewidth]{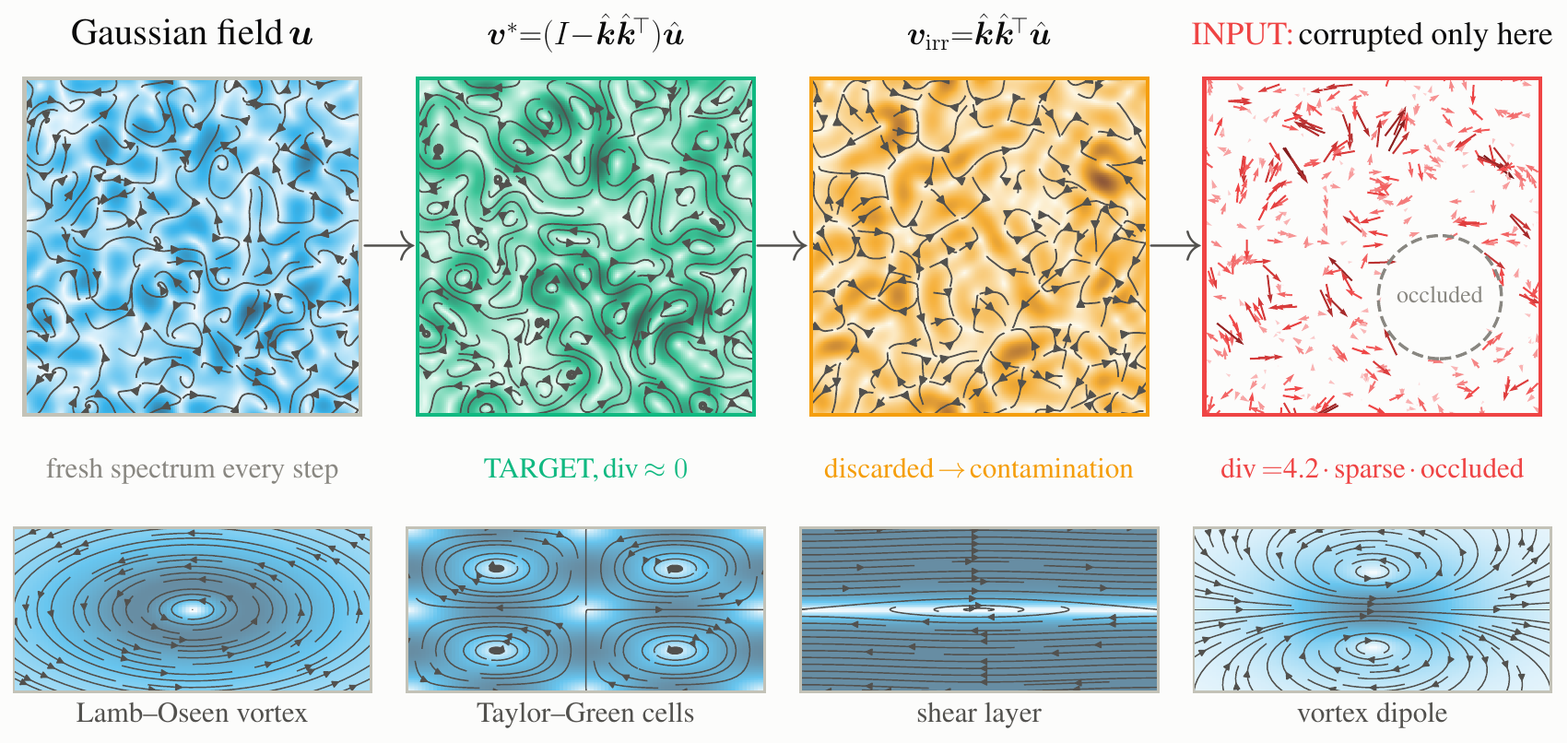}
\caption{\textbf{Training-data pipeline} (2D shown, identical principles in 3D). \emph{Top:} a Gaussian random field splits \emph{exactly} in Fourier space into the solenoidal target $\vstar$ and an irrotational part, sparse observations are sampled and corrupted, and the resulting measurements are used to reconstruct the clean divergence-free target. Panel divergences are measured spectrally. \emph{Bottom:} examples from the CAF library.}
\label{fig:data-pipeline}
\end{figure}

This appendix specifies the data engine behind Section~\ref{sec:data}, summarized in Fig.~\ref{fig:data-pipeline}. Every quoted value is the production setting used for the models of Section~\ref{sec:results}.

\subsection{Conventions: domain, precision, and randomness}
\label{app:data-conventions}

Fields are generated on the periodic unit cube ($L=1$) using a fixed $64^3$ FFT grid. Synthesis uses float64 precision, with float32 batches during training. Data generation is fully deterministic through seeds derived from the training step, sample index, and distributed worker identifier, making the distribution independent of worker count. Training data are generated online, exposing the model to roughly $1.9\times10^{7}$ unique fields throughout training ($4$ workers $\times$ $32$ fields $\times$ $150{,}000$ steps).

\subsection{Primary generator: band-limited Gaussian random fields with an exact Leray split}
\label{app:data-spectral}

We shape three real white-noise grids $w$ in Fourier space, $\widehat{\obsu}=A\,\widehat{w}$, using an
anisotropic von-K\'arm\'an-like spectrum:
\begin{equation}
    A(\vk) = \frac{\sqrt{E(k_{\mathrm{eff}})}}{\max(\|\vk\|,1)},
    \qquad
    E(k) = \frac{(k/k_0)^4}{\bigl(1+(k/k_0)^2\bigr)^{p}},
    \qquad
    k_{\mathrm{eff}}^2 = \sum_{i=1}^{3}\Bigl(\frac{k_i}{a_i}\Bigr)^{2}.
    \label{eq:data-spectrum}
\end{equation}
Here $\vk$ is the integer FFT index (cycles per domain), $a_i$ are per-axis stretch factors, and the
$\|\vk\|$ divisor cancels the $k^2$ growth of modes per shell. As a result, $E$ is the shell energy and
$k_0$ is the energy-containing wavenumber. We zero the DC bin and remove modes with
$\|\vk\|>k_{\mathrm{cut}}G$. The exact Leray projector then splits the field into the target and an
irrotational contamination component:
\begin{equation}
    \widehat{\vstar} = \bigl(I-\hat{\vk}\hat{\vk}^{\!\top}\bigr)\widehat{\obsu},
    \qquad
    \widehat{\vv}_{\mathrm{irr}} = \hat{\vk}\hat{\vk}^{\!\top}\widehat{\obsu},
    \qquad
    \hat{\vk}=\vk/\|\vk\|.
    \label{eq:data-leray}
\end{equation}
Because this is a projection rather than a finite-difference subtraction, $\vstar$ has a spectral
divergence RMS of $\sim\!10^{-14}$ relative to the field RMS. Modes on the Nyquist hyperplanes (axis
index $G/2$) have no conjugate partner, so the projector breaks Hermitian symmetry there. We zero these
modes, which the band limit removes anyway.

Compactly supported divergence-free kernels integrate to zero, so the kernel sum in Eq.~\ref{eq:dfk}
is zero-mean and only $\vmean$ carries net transport. To supervise this term, we add
$\vmean\sim\gN\bigl(\mathbf 0,(0.3\,\mathrm{RMS}(\vstar))^2 I\bigr)$ after the split. The target
remains divergence-free.

\subsection{Capacity-matched band-limiting}
\label{app:data-capacity}

We restrict the training spectra (Table~\ref{tab:data-params}) to scales that the output basis can
represent. $K$ kernels in the unit box have mean spacing $\Delta=L/K^{1/3}$ ($\Delta\approx0.099$ at
$K=1024$). A compactly supported divergence-free basis cannot represent structure much finer than
$\sim\!2\Delta$, so supervision at finer scales adds an irreducible error floor. For example, a fixed
basis reaches $\approx0.9$ relative EPE on broadband spectra, compared with $\approx0.14$ once the
spectra are band-limited to the basis.

The spectral tail, rather than the peak alone, sets the effective length scale, so we constrain all
three spectral parameters:
\begin{itemize}
    \item $k_0\in[1.5,3]$ places the energy-containing wavelength at $3.4$--$6.7$ kernel spacings.
    \item $p\in[3.5,5]$ steepens the roll-off.
    \item $k_{\mathrm{cut}}=0.10$ sets the shortest retained wavelength at $\approx1.6\Delta$, where
    almost no energy remains.
\end{itemize}
We also sample $a_i=1+\gU[-0.3,0.3]$ to prevent the model from learning an isotropy prior.

This design matches supervision to representational capacity. It does not assume that real flows are
band-limited. Accordingly, the synthetic benchmark measures reconstruction within the basis's capacity,
and the real-DNS benchmarks, which are not band-limited, are reported separately.

\subsection{Coherent structures: the CAF library}
\label{app:data-coherent}

Gaussian random fields lack the phase-coherent wakes, jets, shear layers and vortex tubes of real
flows. The second generator superposes $n\sim\gU\{1,\dots,4\}$ closed-form structures from five
families, each with a random center in $[0,L)^3$, random orientation, core radius
$\sim\gU[0.03,0.15]\,L$ and strength $\sim\gU[0.5,2]$, and each divergence-free by a \emph{different}
argument; superposition inherits the property by linearity.

\begin{itemize}
\item \textbf{Lamb--Oseen vortex.} $\vv=v_\theta(r)\,\hat{\ve}_\theta$,
$v_\theta(r)=\frac{\Gamma}{2\pi r}\bigl(1-e^{-r^2/r_c^2}\bigr)$, $r$ the cylindrical radius: any
purely azimuthal field of $r$ alone has $\nabla\!\cdot\!\vv=r^{-1}\partial_\theta v_\theta=0$. The
singularity at $r\to0$ is softened as $r_s=\sqrt{r^2+\epsilon^2}$, a function of $r$ alone, so the
identity survives.
\item \textbf{Taylor--Green cells.} $\vv=A\bigl(\sin kx\cos ky\cos kz,\;-\cos kx\sin ky\cos kz,\;
0\bigr)$, integer $k$: the two nonzero divergence terms cancel term-by-term.
\item \textbf{ABC / Beltrami flow.} $\vv=(A\sin kz+C\cos ky,\;B\sin kx+A\cos kz,\;C\sin ky+B\cos kx)$:
each component depends only on the \emph{other two} coordinates, so every diagonal derivative
vanishes.
\item \textbf{Shear layer.} $\vv=U\tanh\!\bigl((\vx-\vx_0)\!\cdot\!\vn/\delta\bigr)\,\ve$ with $\ve$
Gram--Schmidt-orthogonalized against the layer normal $\vn$ at build time, so
$\nabla\!\cdot\!\vv=f'(s)\,(\vn\!\cdot\!\ve)=0$ exactly, for any profile.
\item \textbf{Vortex ring.} A regularized Biot--Savart sum over $128$ filament segments,
$\frac{\Gamma}{4\pi}\sum_i \mathrm{d}\vl_i\times\vr_i/(\|\vr_i\|^2+a^2)^{3/2}$: each term is the exact
curl of the smooth potential $\mathrm{d}\vl_i/\sqrt{\|\vr_i\|^2+a^2}$, and
$\nabla\!\cdot\!(\nabla\!\times\!\cdot)=0$.
\end{itemize}

These families form the CAF benchmark and are also used to construct the per-family evaluation sets reported in the main paper.

\subsection{Observation corruption}
\label{app:data-corrupt}

Corruption is applied only to the input; the continuous target is never modified.

\textbf{Sampling geometries.} For each field, we draw $M\sim\gU\{M_{\mathrm{lo}},\dots,M_{\mathrm{hi}}\}$ points in $[0,L)^3$ under periodic wrapping, using one of three geometries:
\begin{itemize}
    \item \emph{Uniform:} i.i.d.\ positions.
    \item \emph{Clustered:} half the points fall in $3$--$12$ Gaussian blobs of width
    $\gU[0.03,0.10]\,L$, and the rest are uniform. This mimics patchy Lagrangian measurements.
    \item \emph{Surface:} a randomly oriented plane of thickness $0.02\,L$, mimicking a
    camera-visible sheet.
\end{itemize}

\textbf{Velocity corruption.} At the sampled positions,
\begin{equation}
    \obsum = \vstar(\obspm) + \chi\,\vv_{\mathrm{irr}}(\obspm) + \sigma\,\bm\epsilon_m,
    \qquad
    \bm\epsilon_m\sim\gN(\mathbf 0,I_3),
    \qquad
    \sigma = \nu\,\mathrm{RMS}(\vstar),
    \label{eq:data-corrupt}
\end{equation}
with $\chi\sim\gU[0,\chi_{\max}]$ and $\nu\sim\gU[\nu_{\mathrm{lo}},\nu_{\mathrm{hi}}]$ drawn per
field. For spectral fields, $\vv_{\mathrm{irr}}$ is the target's own Leray complement
(Eq.~\ref{eq:data-leray}), so the divergent contamination is physically consistent with the target
rather than arbitrary noise. The analytic generators have no such complement. For them, $\vv_{\mathrm{irr}}=\nabla\phi$ for a random sum of $8$ low-frequency sinusoids, rescaled to the target RMS.

\textbf{Occlusion.} With probability $p_{\mathrm{occ}}$, we delete all samples inside a ball
occupying a volume fraction $\gU[0.05,0.35]$. Distances use the minimum-image convention, so the
ball wraps correctly across the periodic boundary.

\subsection{Normalization and supervision}
\label{app:data-normalize}

\textbf{Per-field normalization.} We normalize each example before tokenization and invert the transform on output. Positions are mapped to a canonical cube, $\obsp\mapsto(\obsp-\tfrac{L}{2}\mathbf 1)\,(B/L)$ with $B=2$. Velocities are rescaled by $s=1/\mathrm{RMS}(\obsu)$. We compute $s$ from the observed, corrupted input rather than from the target, so the same transform is available at deployment. The target remains in world coordinates:
canonical query points are mapped back, the target is evaluated there, and the result is scaled by the same $s$. The transform is therefore exactly invertible.

\textbf{Supervision.} \label{app:data-supervision}
During training, the field loss is evaluated at $2048$ Monte-Carlo query points per example, resampled at every step, so no location is supervised twice. Evaluation instead uses $4096$ fixed query points per field. Batches are zero-padded to $M_{\max}$ observations. A binary mask excludes padded entries from attention and the losses, and the data loss is normalized per example. Within each curriculum stage, the generator for each example is drawn independently according to that stage's mixture weights (Table~\ref{tab:data-stages}).

\begin{table}[ht!]
\caption{Data-generation parameters. Ranges are drawn independently per field; $\gU$ denotes a uniform
draw (continuous, or over integers where the range is written $\{\cdot\}$). Corruption entries give
the values outside the curriculum; per-stage overrides are in Table~\ref{tab:data-stages}.}
\label{tab:data-params}
\centering
\small
\begin{tabular}{llc}
\toprule
Symbol & Meaning & Value / range \\
\midrule
\multicolumn{3}{l}{\emph{Domain and precision}}\\
$L$                  & periodic domain size, $[0,L)^3$                & $1.0$ \\
$G$                  & FFT generation grid                            & $64^3$ \\
---                  & generation / training precision                & float64 / float32 \\
\midrule
\multicolumn{3}{l}{\emph{Spectral generator (Eq.~\ref{eq:data-spectrum})}}\\
$k_0$                & energy-containing wavenumber (cycles/domain)    & $\gU[1.5,3.0]$ \\
$p$                  & spectral roll-off exponent                     & $\gU[3.5,5.0]$ \\
$a_i$                & per-axis anisotropy stretch                    & $1+\gU[-0.3,0.3]$ \\
$k_{\mathrm{cut}}$   & hard cutoff, $\|\vk\|>k_{\mathrm{cut}}G$        & $0.10$ \\
$\vmean$             & injected constant mean flow                     & $\gN(\mathbf 0,(0.3\,\mathrm{RMS})^2I)$ \\
\midrule
\multicolumn{3}{l}{\emph{Coherent-structure library (CAF)}}\\
$n$                  & superposed structures per field                & $\gU\{1,\dots,4\}$ \\
$r_c$                & core / tube radius                             & $\gU[0.03,0.15]\,L$ \\
---                  & strength ($\Gamma$, amplitude, or speed)        & $\gU[0.5,2.0]$ \\
---                  & ring loop-to-core ratio                        & $\gU[3,8]$ \\
---                  & Taylor--Green / ABC integer wavenumber          & $\gU\{1,2,3\}$ \\
\midrule
\multicolumn{3}{l}{\emph{Corruption (cf. Eq.~\ref{eq:data-corrupt})}}\\
$M$                  & input samples per field                        & $\gU\{256,\dots,4096\}$ \\
---                  & sampling geometry                              & uniform / clustered / surface \\
$\chi$               & irrotational contamination                     & $\gU[0,\chi_{\max}]$ \\
$\nu$                & noise, relative to target RMS                  & $\gU[0,\nu_{\max}]$ \\
$p_{\mathrm{occ}}$   & occlusion probability                          & $0.5$ \\
---                  & occluded volume fraction                       & $\gU[0.05,0.35]$ \\
\midrule
\multicolumn{3}{l}{\emph{Supervision}}\\
---                  & MC query points per example (train)            & $2048$ \\
---                  & frozen query points per field (eval)           & $4096$ \\
---                  & fields per step ($4$ ranks $\times$ $32$)      & $128$ \\
\bottomrule
\end{tabular}
\end{table}

\subsection{Curriculum}
\label{app:data-curriculum}

Three stages change only the data distribution and the loss weights; the model is untouched
(cf. Table~\ref{tab:data-stages}). Core learns the projection itself under mild corruption; Hard, with
heavier contamination and noise, a smaller sample budget, occlusion, and coherent-structure mixing,
is where robustness to heavy corruption and coherent structure is acquired; a short cooldown de-specializes the final
checkpoint from the hardest regime.

\begin{table}[ht!]
    \caption{Training curriculum. Only the data distribution and loss weights change between stages.
    ``$-$'' means the term is off. $\chi_{\max}$ is the maximum irrotational contamination,
    $\nu_{\max}$ the maximum relative noise, $p_{\mathrm{occ}}$ the occlusion probability.}
    \label{tab:data-stages}
    \centering
    \small
    \begin{tabular}{lcccccc} 
        \toprule 
        Stage & Steps & Sources (spectral / coherent) & $\chi_{\max}$ & $\nu_{\max}$ & $M$ & $p_{\mathrm{occ}}$ \\ 
        \midrule 
        Core & $60{,}000$ & $1.0$ / $-$ & $0.5$ & $0.05$ & $[512,4096]$ & $-$ \\ 
        Hard & $85{,}000$ & $0.6$ / $0.4$ & $1.5$ & $0.20$ & $[128,2048]$ & $0.5$ \\ 
        Cooldown & $5{,}000$ & $0.7$ / $0.3$ & $1.0$ & $0.15$ & $[256,4096]$ & $0.5$ \\ 
        \bottomrule 
    \end{tabular} 
\end{table}

\subsection{Time-varying data}
\label{app:data-temporal}

Temporal targets are divergence-free at every instant by construction, not by finite differencing.
We use two families.

\textbf{Modal-spectral (deformation).} We Leray-split $n_{\mathrm{modes}}\sim\gU\{2,3,4\}$ spectral fields $\vv_j$ as in Section~\ref{app:data-spectral} and combine them as
\begin{equation}
    \vv(\vx,\tvar) = \vmean + \sum_j c_j(\tvar)\,\vv_j(\vx), \qquad \tvar\in[0,1],
\end{equation}
where each $c_j$ is a real Fourier series in $\tvar$. The weights do not depend on space, so the sum is divergence-free at every $\tvar$.

\textbf{Advected coherent structures (transport).} We rigidly translate structures from
Section~\ref{app:data-coherent} at constant velocity,
\begin{equation}
    \vv(\vx,\tvar)=\sum_i \vs_i(\vx-\va_i\tvar),
\end{equation}
where each $\va_i$ has a random direction and $\|\va_i\|\sim\gU[0,0.3]$ domain lengths per unit time.
Divergence commutes with spatial shifts, so this construction is exactly divergence-free. We avoid nonlinear self-advection, which would require a pressure projection.

\textbf{Temporal band-limiting.} Following Section~\ref{app:data-capacity}, we match the temporal content of the data to the model's capacity. An order-$F$ temporal expansion represents only harmonics up to $F$, so we use $F_{\mathrm{data}}=9$ harmonics with amplitudes decaying as $(1+f)^{-1}$.

\textbf{Frames, corruption, and supervision.} Each step draws $T\sim\gU\{6,\dots,12\}$ frames at
times $\tvar_i=i/(T-1)$. We corrupt each frame independently (Section~\ref{app:data-corrupt}), which
forces the model to fuse information across time. A single normalization pooled over all frames keeps
the sequence consistent. Query times for the field loss are drawn from $\gU[0,1]$ rather than at the
observed frame times, so the model is also trained to reconstruct at unobserved times.

\subsection{Frozen evaluation sets}
\label{app:data-eval}

Each evaluation set is generated once from a fixed base seed, disjoint from training, with per-field
seed $=$ base $+$ index. Sets are stored as HDF5 files containing the normalized inputs, frozen query
points, targets, and per-field provenance, so every reported number traces back to an exact field.
Synthetic sets use the same band-limited configuration as training
(Section~\ref{app:data-capacity}). The real-DNS sets are not band-limited and are reported
separately. Table~\ref{tab:data-eval} gives the composition of the $582$-flow benchmark.

\begin{table}[t]
\caption{Composition of the $582$-flow evaluation benchmark. Synthetic sets use seeds disjoint from
training. The CAF and single-family sets draw on the training structure library
(Table~\ref{tab:data-stages}), so they measure in-distribution accuracy; held-out families are
evaluated in Table~\ref{tab:res-gen-family}. JHTDB sets are real DNS~\citep{li2008public},
resampled to $64^3$ and used for evaluation only.}
\label{tab:data-eval}
\centering
\small
\begin{tabular}{llcl}
\toprule
Set & Generator & Fields & Role \\
\midrule
in-dist                & spectral, training band-limit           & $128$ & matched-distribution accuracy \\
CAF                    & coherent-structure mixture              & $64$  & coherent structure (in training) \\
Taylor--Green          & single analytic family                  & $64$  & per-family accuracy \\
Lamb--Oseen            & single analytic family                  & $64$  & per-family accuracy \\
vortex ring            & single analytic family                  & $64$  & per-family accuracy \\
shear layer            & single analytic family                  & $64$  & per-family accuracy \\
ABC                    & single analytic family                  & $64$  & per-family accuracy \\
\midrule
isotropic turbulence   & JHTDB (real DNS)                        & $18$  & zero-shot real DNS \\
forced MHD turbulence  & JHTDB (real DNS)                        & $18$  & zero-shot real DNS \\
channel flow           & JHTDB (real DNS)                        & $17$  & zero-shot, wall-bounded \\
transitional b.\ layer & JHTDB (real DNS)                        & $17$  & zero-shot, non-periodic \\
\midrule
\textbf{Total}         &                                         & $\mathbf{582}$ & \\
\bottomrule
\end{tabular}
\end{table}

\subsection{Verification}
\label{app:data-verification}

\textbf{Grid-based targets.} For Leray-split targets, a spectral divergence operator, which is
exact for band-limited periodic fields, gives a divergence RMS of $\sim\!10^{-14}$ relative to the
field RMS.

\textbf{Analytic targets.} These targets have no grid, so we verify them by automatic
differentiation of the continuous field. Relative residuals are $\lesssim10^{-14}$ for all five
families and their superpositions. This holds both at random query points and near vortex-ring
filaments, where the Biot--Savart sum is most strongly cancelling.

\textbf{Temporal targets and inputs.} Temporal targets pass the same checks at multiple times per
sequence. Input divergence is $O(\chi)$ and clearly nonzero
(Fig.~\ref{fig:data-pipeline}), so the reconstruction task is not trivial. Unit tests additionally
cover the normalization round-trip and bit-exact reproducibility across processes.

\section{From the Original DFK Representation to Ours: A Fitting Study}
\label{app:dfk-representation}

We analyze the proposed modifications to the divergence-free kernel (DFK) representation of~\citet{ni2025representing} using \textit{fitting experiments only}. Least-squares fits are performed on clean divergence-free fields without learning, corruption, or sparse observations, making the results properties of the representation itself. The study evaluates four modifications: (i) an explicit mean-flow term, (ii) spacing-relative support radii with strong kernel overlap, (iii) a multi-scale radius mixture, and (iv) anisotropic kernels.

\subsection{Protocol}
\label{app:fit-protocol}
All experiments use a common protocol. Targets comprise band-limited spectral flows (Gaussian Random fields), analytic divergence-free fields, and DNS snapshots from JHTDB~\citep{li2008public}. Coefficients are fitted by ridge least squares, reporting the best ridge value, and evaluated on query points disjoint from those used for fitting. Because the representation is linear in $\{\kwgtk\}$ and $\vmean$ for fixed centers, radii, and shapes, the resulting solution is the optimal approximation within the selected basis. All methods share identical Poisson-disk center placement~\citep{bridson2007poisson}, use float64 precision, and enforce divergence-freeness by construction.

\subsection{A constant mean-flow term}
\label{app:mod-meanflow}

\begin{figure}[!ht]
\centering
\includegraphics[width=0.8\linewidth]{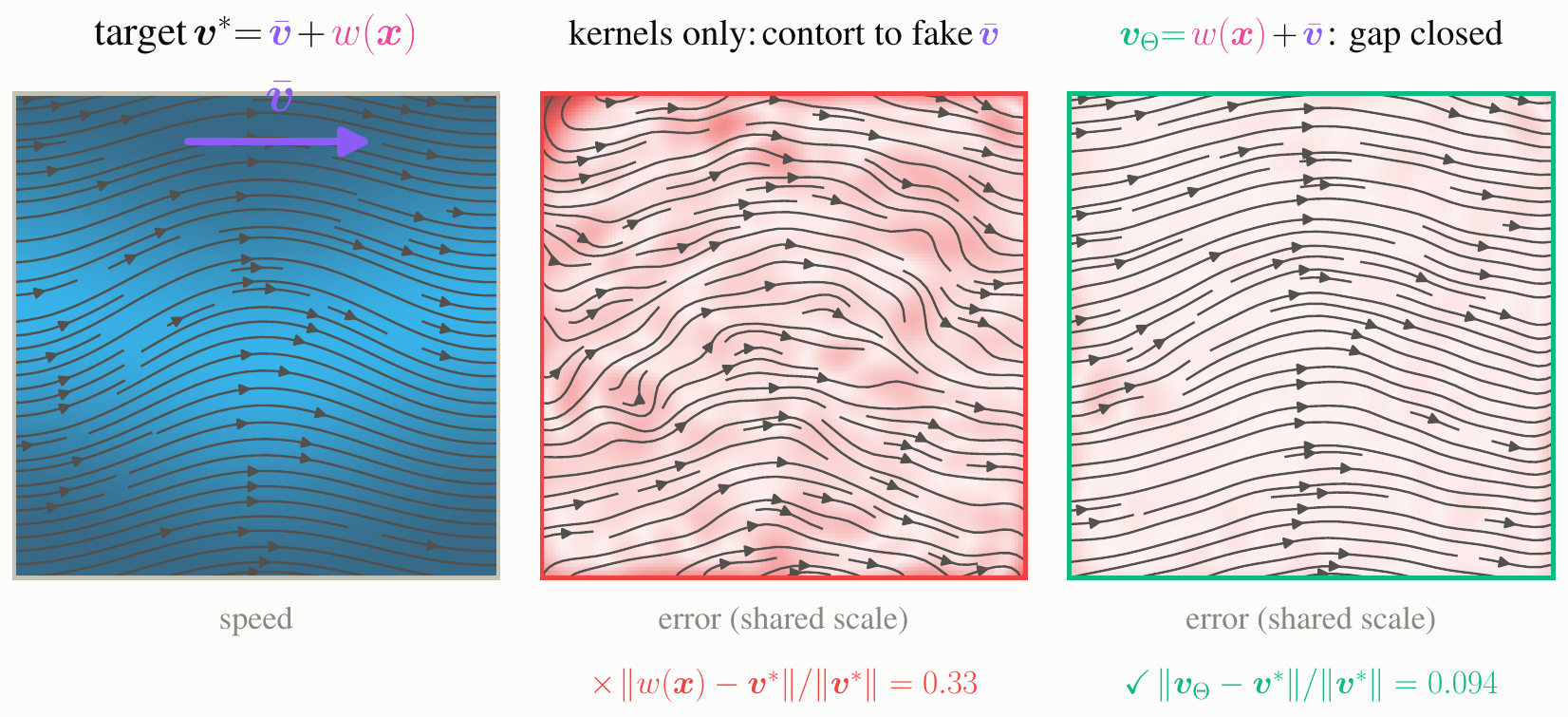}
\caption{\textbf{The mean-flow gap.} Kernels alone cannot represent net transport because their sum is zero-mean. Adding a mean-flow term substantially reduces approximation error.}
\label{fig:meanflow-gap}
\end{figure}

Every compactly supported kernel integrates to zero:
$\int(-\mathbf{I}\nabla^2+\nabla\nabla^{\!\top})\phi\,\mathrm{d}\vz=0$, implying that a finite kernel sum cannot represent a uniform flow component. Since the Leray projector preserves the mean, any target with non-zero transport lies partly outside the original span. Adding a constant velocity term $\vmean$ resolves this limitation.

Figure~\ref{fig:meanflow-gap} illustrates the effect on a flow composed of a uniform component and a zero-mean vortical field. Using kernels alone yields relEPE $0.334$, whereas adding $\vmean$ reduces the error to $0.094$ while accurately recovering the underlying mean velocity. The term is therefore essential for flows with non-zero mean and incurs no cost on zero-mean fields.

\subsection{Support radius: the kernels must overlap heavily}
\label{app:mod-radius}

The packing-based radius of~\citet{ni2025representing} yields supports smaller than the mean kernel spacing $\Delta=(V/K)^{1/3}$, resulting in limited overlap between neighboring kernels. For divergence-free kernels, this regime poorly captures smooth large-scale structures, whereas approximation quality improves substantially as overlap increases. We therefore parameterize the support radius as \[ \kradk = \text{scale}\cdot\Delta, \] and tune the radius scale directly rather than the packing coefficient. Table~\ref{tab:dfk-radius} evaluates the best achievable approximation on real DNS fields at fixed $K=8192$. Increasing overlap consistently lowers the approximation floor across all datasets, often by more than an order of magnitude, indicating that support overlap is a more important design factor than kernel count.
\begin{table}[!h]
\caption{Best achievable fit (relEPE) of the anisotropic DFK span to clean real-DNS
fields~\citep{li2008public} at fixed $K=8192$, as a function of the support radius expressed in kernel
spacings. Overlap is the dominant design variable: the floor falls monotonically as the kernels
overlap more.}
\label{tab:dfk-radius}
\centering
\small
\begin{tabular}{lcccc}
\toprule
Radius / spacing & channel & transitional b.l. & isotropic turb. & forced MHD \\
\midrule
$1.5$                & 0.106 & 0.165 & 0.483 & 0.739 \\
$2.5$                & 0.037 & 0.056 & 0.173 & 0.261 \\
$3.5$                & 0.012 & 0.019 & 0.050 & 0.075 \\
$4.5$                & 0.010 & 0.016 & 0.040 & 0.058 \\
$6.0$                & \textbf{0.009} & \textbf{0.014} & \textbf{0.035} & \textbf{0.051} \\
\bottomrule
\end{tabular}
\end{table}

\paragraph{The opposite failure: over-redundancy.} Although larger radii consistently improve approximation quality, excessive overlap can make neighboring kernels highly redundant, reducing the effective rank of the basis and degrading numerical conditioning. In our proposed model, we employ a multi-scale radius mixture with scales $\{5.0,4.0,3.0\}$ and proportions $\{0.3,0.5,0.2\}$. This choice maintains the strong overlap required to represent smooth large-scale flow structures, while distributing kernels across several support sizes to improve coverage of multiple spatial scales and limit redundancy.

\subsection{Anisotropic kernels}
\label{app:mod-aniso}
The original DFK basis is isotropic, whereas many flow structures such as vortex tubes, shear layers, and jets are elongated. Using the Piola-transformed anisotropic kernel from Eq.~\ref{eq:aniso-kernel} preserves exact divergence-freeness while adapting support to directional structures. Table~\ref{tab:dfk-aniso} compares isotropic and anisotropic kernels on a vortex-tube target using identical center placement, radius scale, and ridge regularization, with orientations derived from local vorticity. Anisotropic kernels consistently improve accuracy at equal kernel count and remain advantageous under equal-memory comparisons, despite requiring additional parameters per kernel. Three conclusions emerge. First, preserving the transverse radius while elongating the kernel (``extend-only'': ($a_\perp=\kradk$, $a_{\mathrm{long}}=\kradk\rho$)) is consistently superior to equal-volume scaling ($a_\perp=\kradk/\rho^{1/3}$, $a_{\mathrm{long}}=\kradk\rho^{2/3}$), which can even underperform the isotropic basis. Second, orientation alignment helps only when coherent directional structure is present; on isotropic spectral flows, aligned and random orientations perform similarly. Third, cylindrical anisotropy provides the best accuracy-memory trade-off, outperforming more expressive triaxial parameterizations once parameter cost is accounted for.

\begin{table}[t]
\caption{Anisotropic vs isotropic fit (relEPE) on a vortex-tube target: identical centers, radius
scale $3.5\,\Delta$, Piola construction, aspect ratio $4$, $8000$ fit samples, $4000$ scoring points,
ridge $10^{-4}$. Same-memory reduces the anisotropic count to $\lfloor 7K/11\rfloor$. The
equal-volume column shows that preserving support volume is worse than not using anisotropy at all.}
\label{tab:dfk-aniso}
\centering
\small
\begin{tabular}{lcccc}
\toprule
& & \multicolumn{2}{c}{same-$K$} & same-memory \\
\cmidrule(lr){3-4}\cmidrule(lr){5-5}
$K$ & isotropic & aniso, equal-volume & aniso, extend-only & aniso, extend-only \\
\midrule
$256$  & 0.173 & 0.251 & \textbf{0.139} & 0.173 \\
$512$  & 0.137 & 0.247 & \textbf{0.092} & \textbf{0.122} \\
$1024$ & 0.128 & 0.282 & \textbf{0.055} & \textbf{0.076} \\
$2048$ & 0.124 & 0.356 & \textbf{0.038} & \textbf{0.045} \\
$4096$ & 0.215 & 0.592 & \textbf{0.049} & \textbf{0.038} \\
\bottomrule
\end{tabular}
\end{table}

\subsection{What fitting alone does \emph{not} justify: adaptive placement}
\label{app:mod-adaptive}

Placement is the primary source of remaining approximation error in the fitting study, motivating the adaptive strategy of Section~\ref{sec:adaptive}. An oracle experiment that optimizes kernel centers and radii directly against the target field yields mixed results: improvements appear on flows requiring precise localization (isotropic turbulence $0.107\rightarrow0.097$, forced MHD
$0.161\rightarrow0.126$), while performance can deteriorate when the fixed basis already provides a good representation (channel $0.025\rightarrow0.043$, transitional boundary layer $0.043\rightarrow0.056$), highlighting the difficulty of the underlying non-convex optimization. Furthermore, adaptive placement increases the parameter count from three kernel weights to seven parameters per kernel, making reliable estimation substantially more data demanding. Together, these results suggest that adaptivity is valuable, but should be learned from data rather than optimized independently for each field.

\subsection{Kernel normalization with per-kernel radii}
\label{app:mod-sigma}

When kernel radii vary, the DFK operator acquires an additional factor of $\kradk^{-2}$ from the chain rule applied to the normalized coordinate $r=\|\vz\|/\kradk$. Divergence-freeness is preserved regardless of this factor, but omitting it alters the relative amplitudes of kernels with different radii. While inconsequential when a single radius is used throughout, correct normalization becomes essential in our setting because the representation combines multiple radius scales (Section~\ref{app:mod-radius}) and predicts radii adaptively for each example. We therefore include the $\kradk^{-2}$ factor so that kernels with different support sizes contribute at consistent physical scales, ensuring comparable parameterizations across both multi-scale and adaptively predicted radii.
\section{Additional Ablation Studies}
\label{app:ablations}

This appendix provides additional ablation studies that are absent from the main paper.

\subsection{Temporal Fourier order}
\label{app:abl-F}

We study the effect of the temporal Fourier order $F$ used to parameterize the kernel coefficients (Section~\ref{sec:temporal-method}). The comparison varies only the value of $F$ ($2$ vs.\ $7$), keeping kernel geometry, training data, and optimization settings fixed. On the $60$-flow real-DNS interpolation benchmark, the lower-order model ($F{=}2$) achieves lower median EPE ($0.114$ vs.\ $0.128$) and mean angular error ($4.8^\circ$ vs.\ $5.7^\circ$). The largest gains occur on wall-bounded flows. This result should be interpreted as a capacity-matching effect rather than evidence that smaller $F$ is universally preferable. On the synthetic validation distribution, which contains stronger high-frequency temporal content, the higher-order model ($F{=}7$) performs better. As in Section~\ref{app:data-capacity}, matching model capacity to the temporal complexity of the data is more important than maximizing expressiveness.

\subsection{Controlled Architecture Matrix} 
\label{app:abl-matrix} 
Table~\ref{tab:abl-matrices} summarizes single-factor ablations on a shared reduced-scale baseline. The results identify the set encoder and kernel representation as the dominant contributors to performance. Replacing attention with DeepSets pooling substantially degrades accuracy, while increasing the number of kernels and enabling adaptive placement both provide clear gains. In contrast, architectural scaling and optimization details have smaller effects. Width and Fourier positional encoding yield modest but consistent improvements when evaluated with a sufficiently long training budget, whereas depth and the orthogonality-loss weight have only minor impact ($\leq1.5\%$). Following these observations, small effects are reported using extended training schedules, while short-budget measurements are interpreted only as directional trends. The kernel-count result ($K{:}~512\!\rightarrow\!2048$) should likewise be viewed as a directional signal; the full-resolution sweep in Section~\ref{sec:ablations} provides a more precise characterization of the accuracy--capacity trade-off.

\begin{table}[!h]
\caption{Controlled architecture matrix: single-factor ablations on a shared reduced-scale base
($K{=}512$, $d_{\mathrm{model}}{=}128$, $4$ blocks, Core stage only), scored on the $128$-field
Gaussian Random Fields set. These are ablation \emph{signal} runs, not paper-scale numbers; small effects
are quoted at the $10\times$ step budget (Section~\ref{app:abl-matrix}). Negative $\Delta$ is an
improvement.}
\label{tab:abl-matrices}
\centering
\small
\begin{tabular}{@{}llcc@{}}
\toprule
Factor & Change & $\Delta$ EPE & Budget \\
\midrule
Set encoder            & attention $\to$ DeepSets pooling      & $+49.1\%$ & $3$k steps \\
Kernel count           & $K{:}~512\to2048$ (no $1024$ point)    & $-12.0\%$ & $3$k steps \\
Adaptive placement     & fixed $\to$ predicted centers/radii   & $-7.4\%$  & $3$k steps \\
Width                  & $d_{\mathrm{model}}{:}~128\to384$      & $-4.8\%$  & $30$k steps \\
Fourier position enc.  & off $\to$ on                          & $-2.4\%$  & $30$k steps \\
Orthogonality loss     & $\lambda_{\mathrm{ortho}}{:}~0.1\to0$   & $+0.04\%$ & $30$k steps \\
\bottomrule
\end{tabular}
\end{table}

\subsection{Single-axis corruption sweep}
\label{app:abl-sweep}

\begin{figure}[ht!]
\centering
\includegraphics[width=0.85\linewidth]{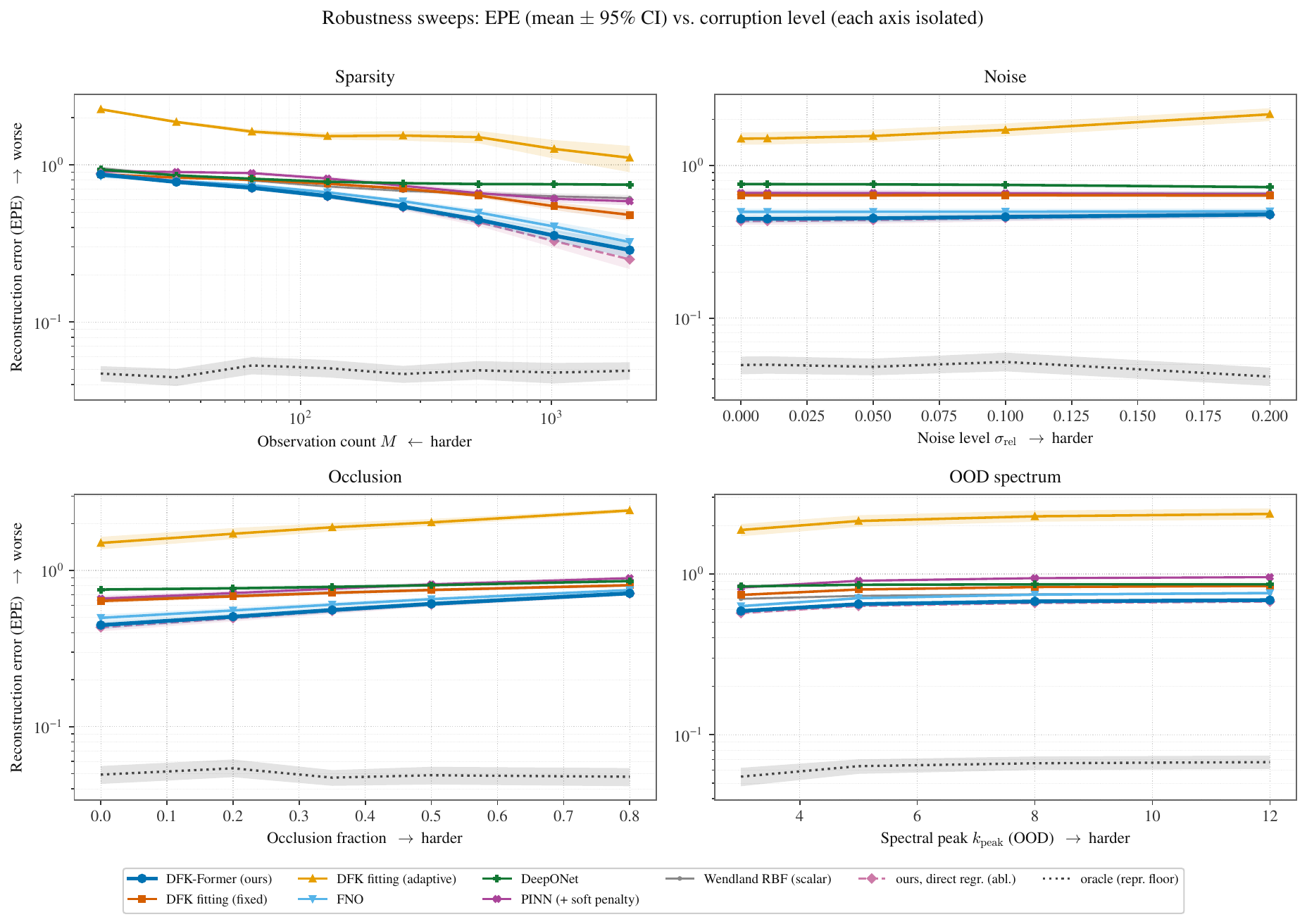}
\caption{\textbf{Axis-isolated robustness sweeps} Reconstruction error (EPE; mean $\pm$ 95\% bootstrap CI; $72$ fields per point) as a function of corruption strength along a single axis, with all other factors fixed to nominal settings. DFK-Former (blue) is the strongest, exactly divergence-free feed-forward method. The dashed pink curve is an ablated variant of DFK-Former, while the grey dotted line denotes the oracle representation floor obtained by least-squares fitting the basis to the clean target, i.e., the best achievable error within the representation.}
\label{fig:abl-robustness}
\end{figure}

We evaluate robustness by varying one corruption factor at a time while keeping the others fixed: sample count, noise level, occluded fraction, and target spectral content. Figure~\ref{fig:abl-robustness} compares multiple baselines across the full range of operating conditions, while Tables~\ref{tab:abl-robust-methods} and \ref{tab:abl-sweep} summarize the hardest settings and a higher-statistics comparison against DFK fitting. Across all operating points, DFK-Former is the strongest feed-forward method that exactly preserves incompressibility. Gains are particularly pronounced over optimization-based DFK fitting with adaptive kernels, while FNO and DeepONet degrade more noticeably under sparse observations, strong corruption, and out-of-distribution spectra. A direct-regression ablation built on the same encoder attains similar accuracy but exhibits substantial divergence leakage, highlighting the role of the DFK representation. The comparison against fixed-kernel DFK fitting reveals a consistent advantage across all corruption axes. The margin narrows under severe corruption, where optimization-based fitting remains relatively robust, but remains largely stable as the target spectrum moves beyond the training range.

\begin{table}[t]
\caption{Multi-method sweep at each axis's hard endpoint ($M{=}64$; $\sigma_{\mathrm{rel}}{=}0.2$ at
$M{=}512$; occluded fraction $0.8$; target $k_{\mathrm{peak}}\in[12,18]$). EPE, lower is better; the
last column is residual divergence. The row before the floor is our own ablation, not a baseline: on
the identical backbone it matches our accuracy and leaks divergence.}
\label{tab:abl-robust-methods}
\centering
\small
\begin{tabular}{@{}lccccc@{}}
\toprule
Method & $M{=}64$ & noise $0.2$ & occl.\ $0.8$ & $k_{\mathrm{peak}}[12,18]$ & $\lVert\nabla\!\cdot\!v\rVert_2$ \\
\midrule
\textbf{DFK-Former (ours)}            & \textbf{0.714} & 0.477 & \textbf{0.713} & 0.687 & $2\text{--}7\times10^{-7}$ \\
FNO                                   & 0.742 & 0.501 & 0.749 & 0.760 & $0.40\text{--}0.67$ \\
DeepONet                              & 0.816 & 0.723 & 0.856 & 0.862 & $\approx0.2$ \\
DFK fitting (fixed kernels)$^\ddag$   & 0.804 & 0.636 & 0.805 & 0.842 & $\sim\!10^{-16}$ \\
DFK fitting (adaptive kernels)        & 1.628 & 2.167 & 2.425 & 2.361 & $\sim\!10^{-15}$ \\
Wendland RBF (scalar)                 & 0.795 & 0.657 & 0.803 & 0.756 & $1.5\text{--}3.1$ \\
PINN ($+$ soft penalty)               & 0.886 & 0.649 & 0.894 & 0.954 & $0.83\text{--}1.5$ \\
\midrule
\multicolumn{6}{@{}l}{\emph{Ablation of our own model, not a baseline}} \\
Set-Transf.\,+\,MLP (direct)\,$^\dag$ & 0.715 & \textbf{0.475} & \textbf{0.713} & \textbf{0.672} & $0.24\text{--}0.76$ \\
\midrule
\emph{oracle representation floor}    & \emph{0.053} & \emph{0.041} & \emph{0.048} & \emph{0.068} & $\sim\!10^{-15}$ \\
\bottomrule
\end{tabular}

\vspace{2pt}
{\footnotesize $^\dag$ Identical encoder, value decoder in place of the DFK head (Section~\ref{sec:ablations}). $^\ddag$ At its per-point best-EPE ridge.}
\end{table}

\begin{table}[ht!]
\caption{Single-axis corruption sweep: model EPE against per-field DFK fitting (fixed kernels) at
its per-point best ridge, $500$ fields $\times$ $3$ seeds per point, $95\%$ bootstrap confidence
intervals non-overlapping at every point. The margin is
$1-\mathrm{EPE}_{\mathrm{model}}/\mathrm{EPE}_{\mathrm{fit}}$; the last column repeats it at the
hard endpoint against a fit over the model's own anisotropic basis.}
\label{tab:abl-sweep}
\centering
\small
\begin{tabular}{@{}llccc@{}}
\toprule
Axis & Range (easy $\to$ hard) & model EPE & margin vs fit & margin vs aniso.\ fit \\
\midrule
Sample count $M$      & $2048\to64$         & $0.315\to0.713$ & $36.3\%\to11.2\%$ & $\to15.8\%$ \\
Noise $\nu$           & $0\to0.4$           & $0.326\to0.390$ & $36.8\%\to26.5\%$ & $\to27.7\%$ \\
Occluded fraction     & $0\to0.5$           & $0.306\to0.492$ & $36.2\%\to26.0\%$ & $\to29.5\%$ \\
Target $k_{\min}$--$k_{\max}$ & $[3,5]\to[12,18]$ & $0.441\to0.544$ & $29.3\%\to32.0\%$ & $\to38.0\%$ \\
\bottomrule
\end{tabular}
\end{table}
\section{Additional Results}
\label{app:results}

In this section, we will present additional results and comparisons to evaluate DFK-Former against the different defined baselines.

\subsection{Per-set breakdown of the main benchmark}
\label{app:res-perset}

Table~\ref{tab:res-perset} reports per-set results for the main benchmark. DFK-Former achieves the lowest EPE on all eleven evaluation sets, with no competing method outperforming it on any column. Interestingly, the real DNS benchmarks are not the most challenging cases: all methods obtain lower errors on the four JHTDB subsets than on the Gaussian Random Fields (GRF) set. For DFK-Former, the average EPE drops from $0.343$ on GRF to $0.153$ across the JHTDB subsets. As discussed in Section~\ref{app:data-capacity}, GRF concentrates energy near the representation's spectral limit, making it a particularly demanding test of approximation capacity. Finally, the two curl-of-potential INRs perform worse than the trivial zero-field predictor (relative $L^2 > 1$) on every synthetic set, indicating that exact incompressibility alone does not provide a sufficiently strong inductive bias for accurate reconstruction.

\begin{table}[ht!]
\caption{\textbf{Per-set breakdown of the main benchmark} (\ensuremath{\EPE}, lower is better).
\textbf{Bold} marks the column minimum. \emph{GRF} is the frozen Gaussian Random Fields set and
\emph{CAF} the coherent-structure mixture (Table~\ref{tab:data-eval}). No method has trained on
JHTDB, so those four columns are zero-shot throughout.}
\label{tab:res-perset}
\centering
\resizebox{\textwidth}{!}{
\begin{tabular}{lccccccccccc}
\toprule
 & \multicolumn{7}{c}{Synthetic (band-limited)} & \multicolumn{4}{c}{Real DNS (zero-shot)} \\
\cmidrule(lr){2-8}\cmidrule(lr){9-12}
Method & GRF & CAF & Taylor--Gr. & Lamb--Os. & vtx.\ ring & shear & ABC
       & iso. & MHD & chan. & tr.\ BL \\
\midrule
Wendland RBF (scalar) & 0.634 & 0.615 & 0.607 & 0.627 & 0.560 & 0.622 & 0.618 & 0.566 & 0.553 & 0.601 & 0.572 \\
SIREN ($+$ soft penalty) & 0.550 & 0.537 & 0.560 & 0.435 & 0.444 & 0.438 & 0.418 & 0.256 & 0.310 & 0.183 & 0.180 \\
Curl-SIREN & 0.998 & 1.177 & 1.167 & 1.102 & 1.028 & 1.108 & 1.005 & 0.673 & 0.729 & 0.769 & 0.699 \\
Vector-potential INR & 0.819 & 1.029 & 0.969 & 0.943 & 0.942 & 0.989 & 0.821 & 0.469 & 0.470 & 0.623 & 0.485 \\
DFK fitting (fixed kernels) & 0.545 & 0.643 & 0.482 & 0.715 & 0.692 & 0.743 & 0.548 & 0.752 & 0.732 & 0.889 & 0.842 \\
DFK fitting (adaptive kernels) & 0.602 & 0.603 & 0.536 & 0.520 & 0.462 & 0.575 & 0.527 & 0.546 & 0.566 & 0.575 & 0.488 \\
FNO & 0.375 & 0.254 & 0.146 & 0.263 & 0.281 & 0.333 & 0.136 & 0.251 & 0.311 & 0.166 & 0.167 \\
FNO ($+$ Leray projection) & 0.372 & 0.270 & 0.139 & 0.262 & 0.296 & 0.420 & 0.129 & 0.246 & 0.305 & 0.163 & 0.162 \\
DeepONet & 0.745 & 0.443 & 0.682 & 0.324 & 0.411 & 0.399 & 0.280 & 0.371 & 0.454 & 0.229 & 0.248 \\
DeepONet ($+$ Leray projection) & 0.744 & 0.456 & 0.681 & 0.347 & 0.429 & 0.486 & 0.278 & 0.366 & 0.451 & 0.223 & 0.242 \\
PINN ($+$ soft penalty) & 0.566 & 0.449 & 0.489 & 0.335 & 0.330 & 0.325 & 0.325 & 0.228 & 0.294 & 0.074 & 0.092 \\
HDNet-3D & 0.391 & 0.196 & 0.143 & 0.146 & 0.188 & 0.165 & 0.105 & 0.266 & 0.301 & 0.201 & 0.216 \\
\textbf{DFK-Former (ours)} & \textbf{0.343} & \textbf{0.101} & \textbf{0.053} & \textbf{0.056} & \textbf{0.062} & \textbf{0.054} & \textbf{0.036} & \textbf{0.200} & \textbf{0.269} & \textbf{0.063} & \textbf{0.081} \\
\bottomrule
\end{tabular}}
\end{table}

\subsection{Enforcing Exact Divergence-Freeness in Operator-Learning Baselines} 
\label{app:res-leray} 
A natural question is whether the advantage of DFK-Former stems primarily from exact incompressibility. To isolate this factor, we post-process the predictions of the operator-learning baselines (FNO and DeepONet), whose raw outputs exhibit non-zero divergence ($0.71$ and $0.23$, respectively), using a standard Leray projection. All experiments are conducted on the same eleven evaluation sets with identical metrics and aggregation procedures.

\paragraph{Leray projection.} 
For each method, the predicted velocity field $\vu$ is represented on a regular $[3,G,G,G]$ grid over the canonical domain, transformed to Fourier space, projected onto the divergence-free subspace, and evaluated at arbitrary query locations via a spectral read-out: 
\begin{equation}
\widehat{\vv}_{\mathrm{sol}}(\vk) = \Bigl(I-\hat{\vk}\hat{\vk}^{\!\top}\Bigr)\widehat{\vu}(\vk), \qquad \vv_{\mathrm{sol}}(\vx) = \frac{1}{G^{3}} \sum_{\vk} \widehat{\vv}_{\mathrm{sol}}(\vk) e^{\,i\,\vk\cdot\vx}, \qquad \hat{\vk}=\vk/\|\vk\|, 
\label{eq:res-leray-readout} 
\end{equation}
using the same angular wavenumbers $2\pi\vk/L$ as Eq.~\ref{eq:data-leray}. Since $\vk\!\cdot\!\widehat{\vv}_{\mathrm{sol}}(\vk)=0$ for every retained Fourier mode, \[ \nabla\!\cdot\!\vv_{\mathrm{sol}}(\vx) = \frac{1}{G^3} \sum_{\vk} i\,(\vk\!\cdot\!\widehat{\vv}_{\mathrm{sol}}(\vk)) e^{\,i\,\vk\cdot\vx} \equiv 0, \] yielding a field that is divergence-free everywhere, not only at grid locations. This property requires spectral evaluation; replacing it with trilinear interpolation breaks the commutativity between projection and reconstruction and increases the divergence norm to $0.23$ for FNO and $0.048$ for DeepONet. FNO is projected directly from its native grid, whereas DeepONet is first evaluated on a $G^3$ grid before projection. The mean-flow component is preserved ($\widehat{P}(\mathbf 0)=I$), Nyquist hyperplanes are removed as in Eq.~\ref{eq:data-leray}, and all computations are performed in \texttt{float64}. 

\paragraph{Exact incompressibility does not recover accuracy.}
The projection incurs only a modest runtime overhead ($1.15\times$ for FNO and $2.15\times$ for DeepONet) and changes macro EPE by $+2.9\%$ and $+2.5\%$, respectively. Despite becoming exactly divergence-free, the projected models remain significantly behind DFK-Former, by factors of $2.10\times$ (FNO) and $3.57\times$ (DeepONet) in macro EPE. 

\paragraph{Projection helps most datasets.}
The aggregate degradation is largely driven by a single failure mode. Across the eleven evaluation sets (Table~\ref{tab:res-perset}), projection improves EPE on $8/11$ datasets for FNO and $7/11$ for DeepONet, including all four real-DNS datasets. The macro-level penalty originates almost entirely from the shear-layer case, where FNO degrades from $0.333$ to $0.420$ and DeepONet from $0.399$ to $0.486$. Excluding this dataset, projection changes macro EPE by only $-0.3\%$ for FNO and $+0.7\%$ for DeepONet. Ablating the pipeline shows that the degradation arises from the projection itself rather than the spectral read-out. For FNO on the shear layer, spectral evaluation alone increases EPE by only $1.0\%$ ($0.333 \rightarrow 0.336$), while the remaining degradation is introduced by the projection. This behavior reflects the periodic-domain assumption of FFT-based solenoidal projection: the shear-layer example contains a strong discontinuity across the periodic boundary, which is interpreted as an irrotational component and consequently removed. In contrast, \textbf{DFK-Former is mesh-free and imposes no periodic-boundary assumption}. 

\paragraph{Grid resolution is not the limiting factor.} 
Since DeepONet requires discretization before projection, we additionally varied the projection grid resolution. Increasing the grid from $G=32$ to $G=64$ changes macro EPE only from $0.4275$ to $0.4253$ ($0.5\%$ improvement) despite an eightfold increase in spectral degrees of freedom. We therefore report results for $G=32$, which also matches FNO's native resolution.

\subsection{Physical consistency: divergence, energy, enstrophy and spectrum}
\label{app:res-physical}

Table~\ref{tab:res-divfree} reveals a clear separation between three classes of methods. Methods that enforce incompressibility \emph{by construction} achieve divergence at numerical precision, including curl-based INRs, DFK fitting, Leray-projected operators, and DFK-Former. Soft-constrained methods reduce divergence but do not eliminate it, as the penalty is enforced only at sampled locations. Unconstrained methods exhibit divergence errors comparable to the scale of the reconstructed field. This separation remains unchanged across all ablations (Section~\ref{app:ablations}).

\begin{table}[hb!]
\caption{\textbf{Divergence quality}, macro-averaged over the eleven static sets; the $L_2$ column is
in Table~\ref{tab:exp-main}. Our $\sim\!10^{-7}$ is the \texttt{float32} autograd
diagnostic's floor, not a property of the field: the DFK construction is solenoidal for any
coefficients, centers and radii (Section~\ref{app:piola_divergence}).}
\label{tab:res-divfree}
\centering
\small
\begin{tabular}{lcc}
\toprule
Method & $\lVert\nabla\!\cdot\!\vfld\rVert_{\infty}$ $\downarrow$ & div-free \\
\midrule
Wendland RBF (scalar) & $1.9{\times}10^{1}$ & no \\
SIREN ($+$ soft penalty) & $7.9{\times}10^{0}$ & no (soft) \\
Curl-SIREN & $5.4{\times}10^{-14}$ & by construction \\
Vector-potential INR & $2.7{\times}10^{-14}$ & by construction \\
DFK fitting (fixed kernels) & $2.8{\times}10^{-14}$ & by construction \\
DFK fitting (adaptive kernels) & $4.3{\times}10^{-14}$ & by construction \\
FNO & $3.5{\times}10^{0}$ & no \\
FNO ($+$ Leray projection) & $5.0{\times}10^{-16}$ & by construction \\
DeepONet & $8.9{\times}10^{-1}$ & no \\
DeepONet ($+$ Leray projection) & $2.5{\times}10^{-16}$ & by construction \\
PINN ($+$ soft penalty) & $2.8{\times}10^{0}$ & no (soft) \\
HDNet-3D & $4.1{\times}10^{0}$ & partially \\
\textbf{DFK-Former (ours)} & $2.6{\times}10^{-6}$ & by construction \\
\bottomrule
\end{tabular}
\end{table}

\begin{table}[ht!]
\caption{\textbf{Physical-consistency battery}, macro-averaged over the three turbulence sets
(Gaussian Random Fields, JHTDB isotropic, JHTDB MHD; $128{+}18{+}18$ fields; spectral curl on the
node-aligned $64^3$ grid). The last two columns repeat enstrophy and $E(k)$ over the ten-family
battery, available for the three methods run over all families.}
\label{tab:res-physical}
\centering
\small
\begin{tabular}{lccc@{\hskip 1.4em}cc}
\toprule
 & \multicolumn{3}{c}{Three turbulence sets} & \multicolumn{2}{c}{Ten-family battery} \\
\cmidrule(lr){2-4}\cmidrule(lr){5-6}
Method & rel.\ KE $\downarrow$ & rel.\ $\Omega$ $\downarrow$ & band $E(k)$ $\downarrow$
       & rel.\ $\Omega$ $\downarrow$ & band $E(k)$ $\downarrow$ \\
\midrule
SIREN ($+$ soft penalty) & \textbf{0.077} & \textbf{0.296} & 0.175 & -- & -- \\
Curl-SIREN & 1.388 & 4.962 & 1.576 & -- & -- \\
DFK fitting (fixed kernels) & 0.148 & 0.486 & 0.230 & -- & -- \\
FNO & 0.202 & 0.696 & 0.258 & 2.660 & 0.211 \\
DeepONet & 0.388 & 0.875 & 0.435 & -- & -- \\
PINN ($+$ soft penalty) & 0.104 & 0.430 & 0.208 & -- & -- \\
HDNet-3D & 0.184 & 0.383 & 0.242 & 1.706 & 0.176 \\
\textbf{DFK-Former (ours)} & 0.144 & 0.470 & \textbf{0.172} & \textbf{0.331} & \textbf{0.073} \\
\bottomrule
\end{tabular}
\end{table}

Beyond divergence, Table~\ref{tab:res-physical} evaluates kinetic energy (KE), enstrophy ($\Omega$), and energy-spectrum fidelity. DFK-Former achieves the lowest band-integrated spectrum error on both the turbulence benchmark and the broader ten-family evaluation, indicating the most faithful recovery of multiscale dynamics. While optimization-based methods with soft divergence penalties can achieve slightly lower KE or enstrophy errors on some turbulence datasets, they do so while violating incompressibility and requiring orders of magnitude higher inference cost. Conversely, exact incompressibility alone does not guarantee physical fidelity: Curl-SIREN is divergence-free by construction yet exhibits large KE, enstrophy, and spectral errors. The differences become more pronounced on the ten-family benchmark. Divergence-leaking operator methods accumulate substantial enstrophy error, particularly on coherent-structure flows, where derivative-based diagnostics are most sensitive. DFK-Former maintains the lowest enstrophy and spectral errors overall, demonstrating that exact incompressibility can be achieved without sacrificing physically meaningful flow statistics.

\begin{figure}[ht!]
\centering
\includegraphics[width=0.85\linewidth]{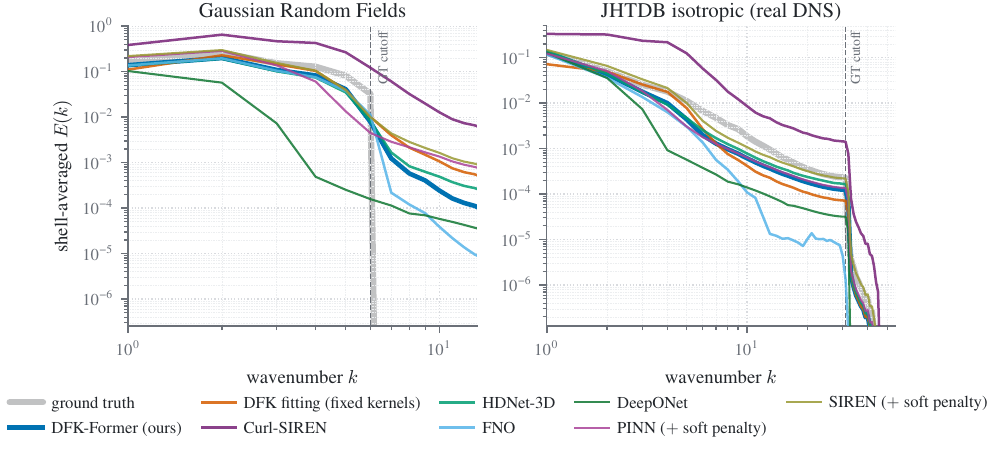}
\caption{\textbf{Energy-spectrum fidelity.} Shell-averaged $E(k)$ (log--log) for ground truth and
every method. Both axes are cropped to the band the ground truth carries: the Gaussian Random Fields
target is band-limited (Section~\ref{app:data-capacity}), so an uncropped axis would compress every curve
into the top of the panel. \mbox{DFK-Former} tracks the ground truth across the resolved band.}
\label{fig:res-spectrum}
\end{figure}

\subsection{Efficiency}
\label{app:res-efficiency}

Table~\ref{tab:res-efficiency} compares inference cost across methods. DFK-Former reconstructs a scene in $7.5$\,ms, achieving the highest throughput ($134$ scenes/s) while requiring the least memory ($68$\,MB). At $M{\approx}2$k observations and $Q{=}256$ queries, a forward pass requires $41$ GFLOPs, approximately half the cost of fixed-kernel DFK fitting ($81$ GFLOPs) and substantially less than optimization-based alternatives. For this experiment, FLOPs were measured with \texttt{torch.utils.flop\_counter} ($2\times$MAC convention) for network-based methods and computed analytically for fixed-kernel DFK fitting.

The computational profile is dominated by the transformer encoder, which accounts for more than $99.9\%$ of the total FLOPs; kernel evaluation is comparatively negligible. As a result, increasing the number of kernels has little impact on runtime over the range considered in Section~\ref{tab:exp-ablation-K}. In contrast, optimization-based methods incur a fundamentally different inference cost, solving a reconstruction problem independently for each field. DFK-Former shifts this cost to training, replacing per-instance optimization with a single feed-forward prediction.

\begin{table}[ht!]
\caption{\textbf{Efficiency battery} Runtime, throughput, peak memory, and computational cost for reconstructing a single scene ($M{\approx}2$k, $Q{=}256$, batch size $1$). Lower is better except for throughput.}
\label{tab:res-efficiency}
\centering
\small
\resizebox{\textwidth}{!}{
\begin{tabular}{lccccc}
\toprule
Method & ms/field $\downarrow$ & scenes/s $\uparrow$ & peak mem (MB) $\downarrow$ & GFLOPs $\downarrow$ & \# params (M) \\
\midrule
SIREN / Curl-SIREN (per-field fit)  & 53000 & 0.019 & 204 & 9500 & 0.20 \\
DFK fitting (fixed kernels)         & 22 & 45 & 1285 & 81 & 0 \\
DFK fitting (adaptive kernels)      & 9821 & 0.10 & 623 & 786 & 0 \\
\textbf{DFK-Former (ours)}          & \textbf{7.5} & \textbf{134} & \textbf{68} & \textbf{41} & 10.2 \\
\bottomrule
\end{tabular}}
\end{table}

\subsection{Generalization}
\label{app:res-generalization}

We evaluate generalization along four axes: unseen sparsity, unseen noise, unseen flow families, and zero-shot transfer to real DNS data. For unseen sparsity and noise, DFK-Former, FNO, and DeepONet are retrained under matched restricted training distributions and evaluated outside the training range (Table~\ref{tab:res-gen-corruption}). For unseen families, training is restricted to Gaussian-random-field and ABC flows, with all remaining coherent flow families withheld (Table~\ref{tab:res-gen-family}). Real-DNS transfer requires no additional training, as none of the methods observe JHTDB data during training.

\paragraph{Corruption-level extrapolation.} Table~\ref{tab:res-gen-corruption} shows that DFK-Former generalizes well beyond the sparsity and noise levels observed during training. Restricting training to denser observations increases error by only $0.5\%$ at the most sparse evaluation point ($M=64$), while restricting training to $\sigma_{\rm rel}\le10\%$ noise increases error by $3.1\%$ when evaluated at $20\%$ noise. Similar behavior is observed for restricted FNO and DeepONet models, indicating that all three architectures are relatively insensitive to moderate extrapolation along these corruption axes. Compared with the matched restricted baselines, DFK-Former consistently achieves the lowest error across sparsity levels and remains the strongest method under severe noise. The relative ranking changes little compared to the full-range models, suggesting that performance differences are not driven by exposure to the test corruption levels during training.

\begin{table}[hb!]
\caption{\textbf{Generalization to unseen sparsity and noise levels.} Restricted models are trained only on moderate sparsity ($M\!\in\![256,1024]$) or noise ($\sigma_{\rm rel}\!\le\!10\%$) and evaluated both within and beyond that range. Columns marked by $^\dagger$ correspond to out-of-distribution settings not observed during training. DFK-Former remains competitive under both sparsity and noise extrapolation, with only modest degradation relative to its full-range counterpart. FNO and DeepONet use identical restricted-training protocols. Results report $\EPE$ averaged over $24$ fields and $3$ random seeds. Lower is better.}
\label{tab:res-gen-corruption}
\centering
\resizebox{\textwidth}{!}{
\begin{tabular}{l ccccc @{\hskip 1.2em} cccc}
\toprule
 & \multicolumn{5}{c}{Sparsity axis} & \multicolumn{4}{c}{Noise axis ($\sigma_{\rm rel}$)} \\
\cmidrule(lr){2-6}\cmidrule(lr){7-10}
Method & $1\%$ ($M{=}64$)$^\dagger$ & $2\%$ ($M{=}128$)$^\dagger$ & $5\%$ & $20\%$ & $50\%$
       & $0\%$ & $5\%$ & $10\%$ & $20\%^\dagger$ \\
\midrule
\multicolumn{10}{l}{\emph{Restricted training} --- sparsity arm $M\!\in\![256,1024]$, noise arm $\sigma_{\rm rel}\!\le\!10\%$, every stage, retrained from step~$0$}\\
FNO                                    & 0.753 & 0.665 & 0.580 & 0.407 & 0.362 & 0.495 & 0.496 & 0.497 & 0.501 \\
DeepONet                               & 0.812 & 0.782 & 0.766 & 0.756 & 0.750 & 0.757 & 0.755 & 0.748 & 0.724 \\
\textbf{DFK-Former (ours)}             & \textbf{0.718} & \textbf{0.637} & 0.551 & 0.380 & 0.323 & 0.461 & 0.464 & 0.471 & \textbf{0.492} \\
\midrule
\multicolumn{10}{l}{\emph{Full-range training} ($M\!\in\![128,2048]$, $\sigma_{\rm rel}\!\le\!20\%$) --- unrestricted reference}\\
FNO                                    & 0.742 & 0.669 & 0.587 & 0.405 & 0.323 & 0.498 & 0.498 & 0.498 & 0.501 \\
DeepONet                               & 0.816 & 0.781 & 0.765 & 0.753 & 0.748 & 0.756 & 0.753 & 0.747 & 0.723 \\
DFK-Former                             & 0.714 & 0.634 & 0.542 & 0.355 & 0.287 & 0.447 & 0.451 & 0.461 & 0.478 \\
\midrule
DFK fitting (fixed kernels, best ridge) & 0.804 & 0.759 & 0.706 & 0.546 & 0.481 & 0.638 & 0.638 & 0.640 & 0.636 \\
\emph{oracle representation floor}     & \emph{0.053} & \emph{0.051} & \emph{0.047} & \emph{0.048} & \emph{0.049} & \emph{0.049} & \emph{0.048} & \emph{0.052} & \emph{0.041} \\
\bottomrule
\end{tabular}}
\end{table}

\begin{table}[ht!]
\caption{\textbf{Flow-family generalization under training-set restriction.} $\EPE$ on the eleven evaluation sets. The \emph{restricted} model is trained using only the ABC coherent-flow family, while keeping all other training components, sampling weights, corruption ranges, and spectral-flow data identical to the full setup. The ratio reports \textit{restricted/full}. Large ratios on withheld families indicate strong transfer from training on diverse coherent structures, whereas ratios near one indicate limited dependence on flow-family diversity. The JHTDB datasets are zero-shot for both models.}
\label{tab:res-gen-family}
\centering
\resizebox{0.9\textwidth}{!}{
\begin{tabular}{llccc}
\toprule
 & Set & restricted & full library & ratio \\
\midrule
\multirow{4}{*}{\textbf{Excluded families}}
 & Shear layer               & 0.425 & 0.054 & \textbf{7.95} \\
 & Taylor--Green             & 0.301 & 0.053 & 5.71 \\
 & Vortex ring               & 0.277 & 0.062 & 4.50 \\
 & Lamb--Oseen               & 0.249 & 0.056 & 4.48 \\
\midrule
\multirow{2}{*}{Included families}
 & ABC (the one kept family) & 0.074 & 0.036 & 2.04 \\
 & Gaussian Random Fields    & 0.369 & 0.343 & \textbf{1.08} \\
\midrule
Representative mixture & CAF ($4/5$ Excluded families)    & 0.307 & 0.101 & 3.04 \\
\midrule
\multirow{4}{*}{\shortstack[l]{Real DNS\\(zero-shot)}}
 & JHTDB channel (wall shear)     & 0.146 & 0.063 & \textbf{2.33} \\
 & JHTDB transitional BL          & 0.157 & 0.081 & 1.93 \\
 & JHTDB isotropic (homogeneous)  & 0.224 & 0.200 & 1.12 \\
 & JHTDB MHD (homogeneous)        & 0.272 & 0.269 & \textbf{1.01} \\
\midrule
 & macro (all eleven sets)   & 0.255 & 0.120 & \textbf{2.13} \\
\bottomrule
\end{tabular}}
\end{table}

\paragraph{Structural generalization is substantially harder.} A markedly different trend emerges when entire flow families are withheld during training (Table~\ref{tab:res-gen-family}). Removing four of the five coherent-flow families increases error by $4.5$--$8.0\times$ on the withheld families and by $2.13\times$ overall. The degradation extends beyond the unseen families themselves: performance on the retained ABC family also deteriorates, whereas the Gaussian Random Fields source distribution remains largely unchanged. These results indicate that coherent-flow diversity contributes directly to the learned prior rather than merely improving interpolation within individual families. 

The effect propagates in a physically meaningful manner. Wall-bounded DNS datasets exhibit the largest degradation, whereas homogeneous turbulence datasets remain comparatively stable. Notably, the restricted model retains the defining properties of the representation, remaining exactly divergence-free and preserving the same inference cost. The loss is therefore attributable to reduced prior coverage rather than a breakdown of the reconstruction mechanism itself.

\subsection{Time-varying flow: the full breakdown}
\label{app:res-temporal}

Tables~\ref{tab:res-temporal}--\ref{tab:res-temporal-variants} provide a detailed evaluation on the dynamic JHTDB benchmark (60 flows, $180$ test frames), including performance across corruption regimes, DNS datasets, and temporal-model variants.

\begin{table}[!bp]
\caption{\textbf{Time-varying reconstruction of unobserved real DNS frames}, all six corruption
regimes (dynamic JHTDB: $60$ flows, $180$ test observations). Each cell is \textbf{median
(mean)} endpoint error at times the model never sees. Per-frame DFK's mean is inflated by $15/180$
numerical failures.}
\label{tab:res-temporal}
\centering
\setlength{\tabcolsep}{4pt}
\resizebox{\textwidth}{!}{
\begin{tabular}{lccccc}
\toprule
Corruption regime & \textbf{Temporal DFK-Former} & Per-frame DFK & Shared-kernel DFK & ST-SIREN & ST-Curl-SIREN \\
\midrule
\texttt{clean\&dense} & 0.114 (0.114) & 0.100 (0.102) & \textbf{0.083} (0.093) & 0.442 (0.432) & 0.781 (0.727) \\
\texttt{contaminated} & \textbf{0.085} (0.122) & 0.217 (0.291) & 0.165 (0.191) & 0.365 (0.448) & 0.712 (0.781) \\
\texttt{noisy} & \textbf{0.088} (0.097) & 0.398 (1.874) & 0.268 (0.281) & 0.440 (0.482) & 0.939 (0.926) \\
\texttt{sparse} & \textbf{0.109} (0.129) & 0.847 (0.828) & 1.093 (1.067) & 0.388 (0.453) & 0.590 (0.617) \\
\texttt{occluded} & \textbf{0.098} (0.110) & 0.951 (107.1) & 1.511 (1.519) & 0.413 (0.451) & 0.766 (0.819) \\
\texttt{hard combo} & \textbf{0.138} (0.173) & 0.916 (11.52) & 1.420 (1.419) & 0.497 (0.528) & 1.029 (1.057) \\
\midrule
\textbf{Overall} & \textbf{0.114} (0.124) & 0.767 (20.29) & 0.626 (0.762) & 0.428 (0.466) & 0.842 (0.821) \\
\bottomrule
\end{tabular}}
\end{table}

\paragraph{Temporal priors matter primarily in ill-posed regimes.} Table~\ref{tab:res-temporal} shows that Temporal DFK-Former achieves the lowest error in every under-constrained setting, including contamination, noise, sparsity, occlusion, and their combination. Relative to per-frame DFK fitting, the improvement ranges from $2.6\times$ to $9.7\times$, with the largest gains under sparse and occluded observations. In contrast, the only regime where the temporal model does not achieve the lowest error is \texttt{clean\&dense}, where observations are sufficiently informative for optimization-based reconstruction to approach the optimal solution within the fixed kernel basis. This contrast highlights the role of the learned temporal prior: its benefits emerge when the reconstruction problem is ambiguous rather than fully determined by the observations. Per-frame DFK fitting additionally exhibits occasional numerical failures, substantially affecting mean errors in the most challenging settings, whereas all learned temporal models remain stable across the entire benchmark.

\paragraph{Benefits increase with temporal coherence.} Table~\ref{tab:res-temporal-bydataset} breaks results down by flow type. Improvements over per-frame reconstruction are largest for wall-bounded flows, reaching $14.2\times$ on channel flow and $10.4\times$ on the transitional boundary layer, while homogeneous turbulence datasets show smaller but still substantial gains. These trends are consistent with the amount of exploitable temporal structure available in each dataset. Across all datasets, the predicted fields remain exactly divergence-free up to the numerical precision of the diagnostic.

\begin{table}[ht!]
\caption{\textbf{Temporal reconstruction by flow type.} Median metrics on the dynamic JHTDB benchmark ($180$ test observations). Temporal DFK-Former consistently outperforms per-frame DFK fitting, with the largest gains on wall-bounded flows. The ratio column reports the relative improvement over a per-frame DFK baseline using the same kernel basis.}
\label{tab:res-temporal-bydataset}
\centering
\small
\begin{tabular}{lcccccc}
\toprule
Dataset & $n_{\mathrm{obs}}$ & \ensuremath{\EPE} & angle ($^\circ$) & $\lVert\nabla\!\cdot\!\vfld\rVert_{\mathrm{rms}}$ & Per-frame DFK & ratio \\
\midrule
JHTDB isotropic & 48 & 0.141 & 6.8 & $5.6{\times}10^{-7}$ & 0.752 & 5.3$\times$ \\
JHTDB MHD & 48 & 0.196 & 9.5 & $6.3{\times}10^{-7}$ & 0.790 & 4.0$\times$ \\
JHTDB channel & 42 & \textbf{0.057} & \textbf{2.7} & $2.1{\times}10^{-7}$ & 0.805 & \textbf{14.2}$\times$ \\
JHTDB transitional BL & 42 & 0.073 & 3.4 & $2.3{\times}10^{-7}$ & 0.753 & 10.4$\times$ \\
\bottomrule
\end{tabular}
\end{table}

\paragraph{Shared anisotropic kernels provide the strongest temporal model.} Table~\ref{tab:res-temporal-variants} compares alternative temporal parameterizations under a fixed kernel budget. The best-performing configuration uses a shared anisotropic kernel geometry across time, while predicting only time-varying kernel coefficients and mean flow. This design outperforms both per-frame optimization and moving-kernel variants, achieving the lowest overall median error ($0.114$). 

Interestingly, the adaptive kernel placement strategy that improves the static model does not transfer to the temporal setting. Although adaptive placement improves synthetic validation performance, it consistently underperforms fixed shared placement on real DNS data. This suggests that a stable spatial basis shared across time provides a stronger inductive bias than frame-specific geometric adaptation. 

\begin{table}[hb!]
\caption{\textbf{Ablation of temporal parameterizations on real DNS sequences.} Median EPE on the dynamic JHTDB benchmark ($60$ flows, $180$ test observations). All temporal variants share the same $K{=}1024$ kernel budget, training data, and supervision. The best-performing design uses a shared anisotropic kernel geometry with low-order temporal dynamics ($F{=}2$), outperforming both moving-kernel variants and per-frame DFK fitting. Adding adaptive kernel placement provides no benefit on real DNS data. Lower is better. }
\label{tab:res-temporal-variants}
\centering
\setlength{\tabcolsep}{3.5pt}
\resizebox{\textwidth}{!}{
\begin{tabular}{lcccccc}
\toprule
Corruption regime & Per-frame DFK & $F{=}2$ moving iso. & $F{=}7$ moving iso. & $F{=}7$ shared aniso. & $F{=}2$ shared aniso. & \,$+$ adaptive placement \\
\midrule
\texttt{clean\&dense}  & 0.100 & 0.115 & 0.174 & 0.133 & \textbf{0.114} & 0.117 \\
\texttt{contaminated}  & 0.217 & 0.092 & 0.172 & 0.105 & \textbf{0.085} & 0.100 \\
\texttt{noisy}         & 0.398 & 0.093 & 0.173 & 0.102 & \textbf{0.089} & 0.098 \\
\texttt{hard combo}   & 0.916 & 0.146 & 0.222 & 0.152 & \textbf{0.138} & 0.152 \\
\texttt{sparse}        & 0.847 & 0.118 & 0.194 & 0.125 & \textbf{0.109} & 0.117 \\
\texttt{occluded}      & 0.951 & 0.103 & 0.186 & 0.117 & \textbf{0.098} & 0.108 \\
\midrule
\textbf{Overall}       & 0.767 & 0.117 & 0.183 & 0.128 & \textbf{0.114} & 0.118 \\
\midrule
angle / $\lVert\nabla\!\cdot\!\vfld\rVert_{\mathrm{rms}}$
 & $40.0^\circ$ / $\sim\!10^{-13}$ & $5.2^\circ$ / $6.4{\times}10^{-7}$
 & $7.6^\circ$ / $6.0{\times}10^{-7}$ & $5.7^\circ$ / $4.9{\times}10^{-7}$
 & $\mathbf{4.8^\circ}$ / $4.6{\times}10^{-7}$ & $5.0^\circ$ / $4.7{\times}10^{-7}$ \\
\bottomrule
\end{tabular}}
\end{table}

\begin{figure}[ht!]
\centering
\includegraphics[width=\linewidth]{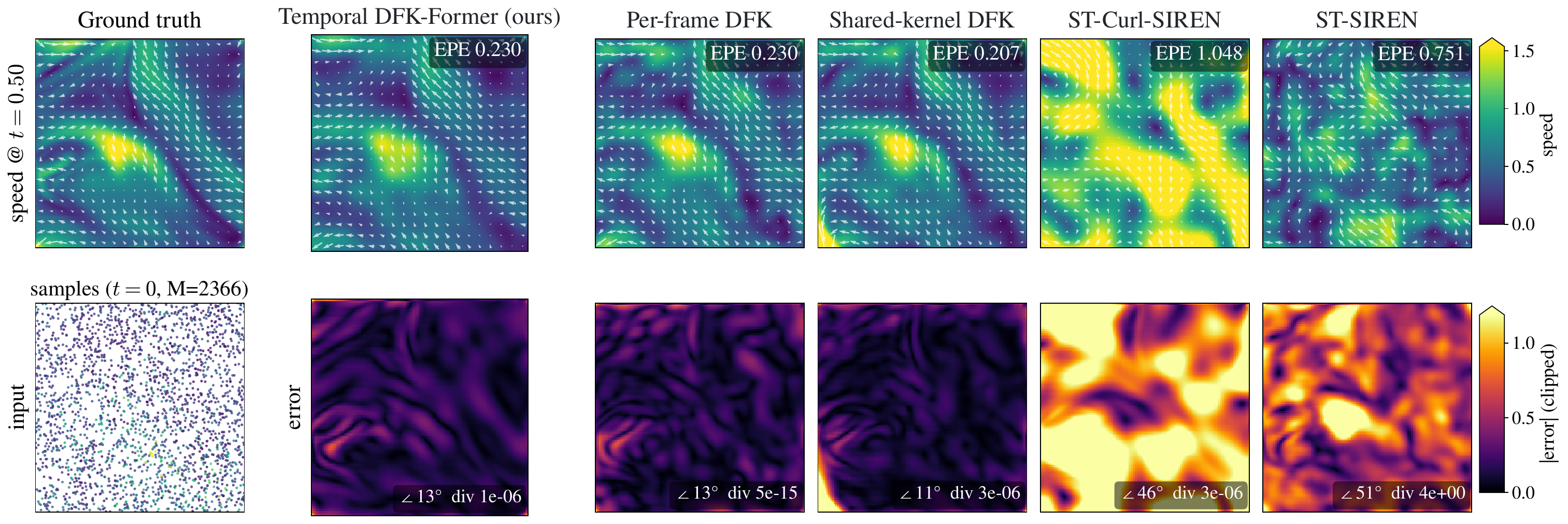}
\caption{\textbf{The well-posed case the model does not win} (Forced MHD (dynamic JHTDB), \texttt{clean\&dense}): the DFK-span solves reach the fixed-basis optimum ---
shared-kernel DFK edging the learned prior out ($0.207$) and per-frame DFK matching it ($0.230$
each). The space-time INRs remain far behind ($0.751$, $1.048$). The learned prior earns its keep in
the ill-posed regimes, not everywhere.}
\label{fig:res-temporal-cleandense}
\end{figure}

\begin{figure}[!h]
\centering
\includegraphics[width=\linewidth]{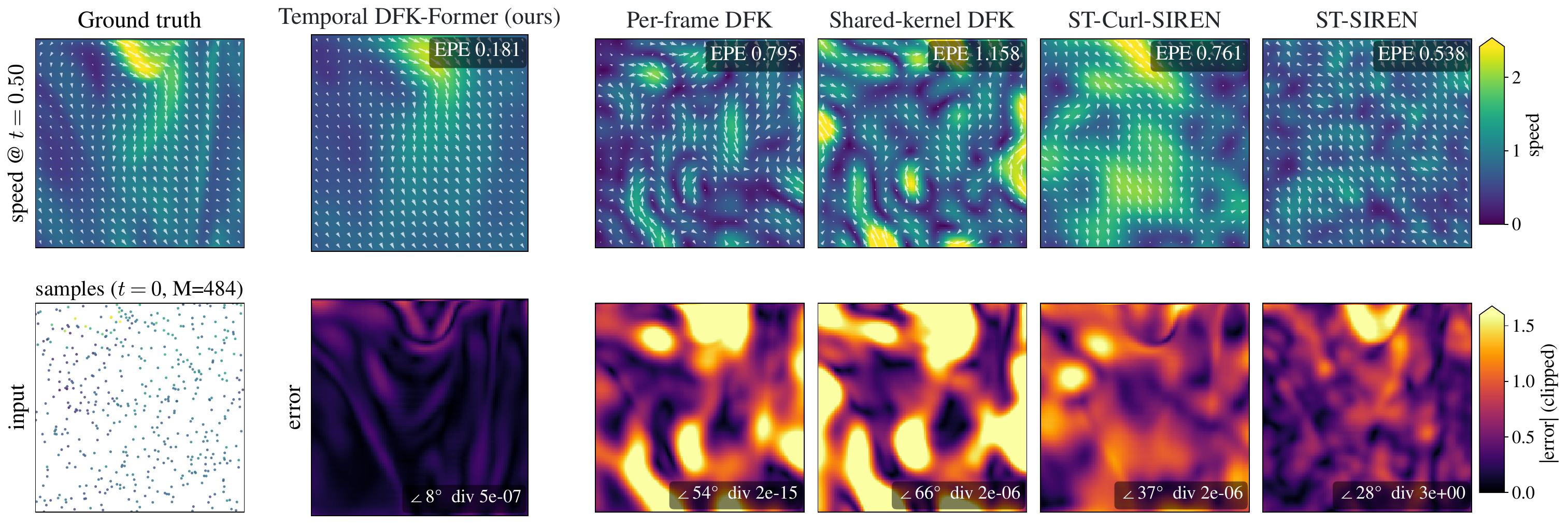}
\caption{\textbf{Temporal reconstruction on a representative real-DNS example} (Forced MHD (dynamic JHTDB), \texttt{sparse} regime). Top: ground truth and predictions from Temporal DFK-Former, per-frame DFK fitting, iterative DFK fitting, space-time Curl-SIREN, and space-time SIREN. Bottom: sparse observations and reconstruction errors. DFK-Former best preserves the flow structure and achieves the lowest EPE while remaining exactly divergence-free.}
\label{fig:temporal-baselines}
\end{figure}

\paragraph{Qualitative comparison across observation regimes.} Figures~\ref{fig:res-temporal-cleandense} and~\ref{fig:temporal-baselines} compare representative Forced MHD sequences under two contrasting observation regimes: a well-observed Forced MHD \texttt{clean\&dense} setting ($M\!\approx\!2400$ samples per frame) and a severely under-constrained \texttt{sparse} setting ($M\!\approx\!420$). In both figures, the top row shows the ground truth and reconstructed velocity magnitude at an unseen time instant, while the bottom row visualizes the corresponding observations and error maps. 

In the Forced MHD \texttt{clean\&dense} regime (Figure~\ref{fig:res-temporal-cleandense}), optimization-based DFK reconstruction attains the lowest errors. Shared-kernel DFK achieves the best result within the DFK basis, and per-frame DFK closely matches it, while Temporal DFK-Former reaches comparable accuracy. The resulting error maps are nearly indistinguishable, indicating that the observations are sufficiently informative for optimization alone to recover the best solution supported by the representation. In this regime, the primary advantage of the learned model is computational efficiency rather than reconstruction accuracy. 

The behavior changes markedly in the \texttt{sparse} regime (Figure~\ref{fig:temporal-baselines}), where the observations no longer uniquely determine the flow. Both optimization-based DFK variants deteriorate substantially, with errors concentrated in regions far from the observed samples. In contrast, Temporal DFK-Former preserves the dominant flow structures and maintains low reconstruction error despite the reduced observation density. This qualitative trend mirrors the aggregate results in Table~\ref{tab:res-temporal}: between the Forced MHD \texttt{clean\&dense} and \texttt{sparse} regimes, the median error of per-frame DFK increases from $0.100$ to $0.847$, while Temporal DFK-Former remains essentially unchanged ($0.114$ to $0.109$). 

The space-time SIREN baseline exhibits visually plausible reconstructions in some sparse examples, but does not satisfy the incompressibility constraint, exhibiting divergence residuals several orders of magnitude larger than the kernel-based methods. Taken together, these examples illustrate the complementary strengths of optimization and learning: optimization-based fitting is most effective when observations densely constrain the solution, whereas the learned temporal prior provides the largest benefit when reconstruction is under-constrained.

\begin{figure}[ht!]
\centering
\includegraphics[width=0.55\linewidth]{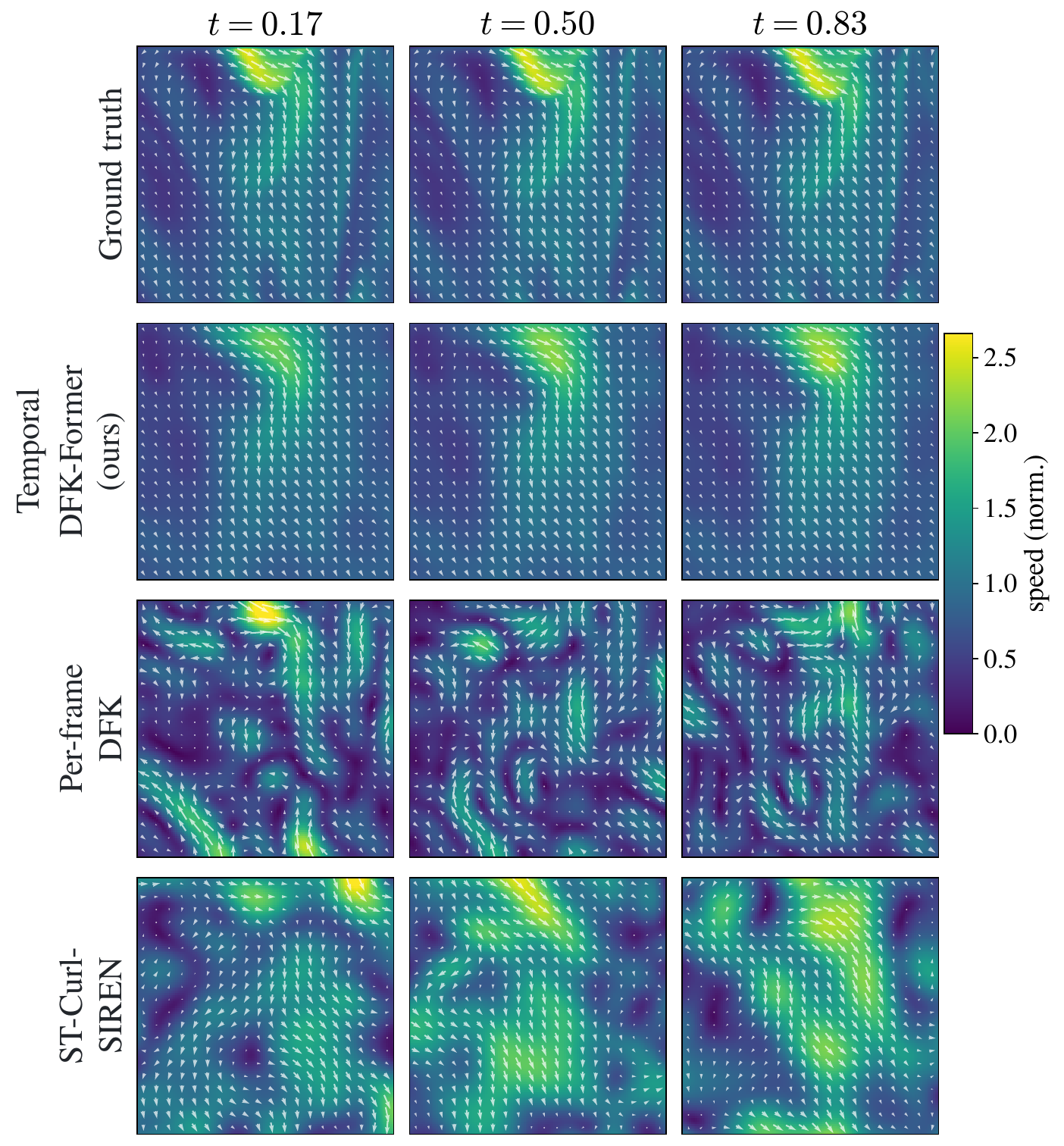}
\caption{\textbf{The reconstruction as a function of time} (Forced MHD (dynamic JHTDB), \texttt{sparse} regime). Columns are three real-DNS times the model never observes ---
frame midpoints; rows are ground truth, our model, and two baselines. The emitted field is continuous
in $\tvar$ and exactly divergence-free at every $\tvar$, not only where it was supervised.}
\label{fig:res-temporal-strip}
\end{figure}

Figure~\ref{fig:res-temporal-strip} further illustrates representative predictions across time. The model produces a continuous velocity field over $\tvar$ while preserving exact divergence-freeness at unseen intermediate time instances.

\end{document}